\documentclass[a4paper,fleqn]{cas-dc}

\usepackage[T1]{fontenc}
\usepackage[english]{babel}
\usepackage{tgbonum}
\usepackage{amsmath,amssymb,amsthm,mathrsfs,bm}
\usepackage{graphicx}
\usepackage{booktabs}
\usepackage{multirow}
\usepackage{tabularx}
\usepackage{siunitx}
\usepackage[table]{xcolor}
\usepackage{float}
\usepackage{subcaption}
\usepackage{makecell}
\usepackage{enumitem}
\usepackage[ruled,vlined]{algorithm2e}
\usepackage{orcidlink}

\usepackage{xurl}
\usepackage[labelfont=,labelsep=period]{caption}
\begin{document}

\let\WriteBookmarks\relax
\def\floatpagepagefraction{1}
\def\textpagefraction{.001}

\shorttitle{DVL-DeepONet}
\shortauthors{Sahoo and Klein}

\title[mode=title]{DVL-DeepONet: A Physics-Guided Operator Learning for Resilient Underwater Navigation}

\author[1]{Arup Kumar Sahoo} [
orcid=0000-0003-4515-7434
] \credit{Conceptualization, Methodology, Software, Validation, Investigation, Writing - original draft}
\cormark[1]
\ead{asahoo@campus.haifa.ac.il}

\author[1]{Itzik Klein} [
orcid=0000-0001-7846-0654
] \credit{Supervision, Methodology, Writing - review \& editing}
\ead{kitzik@univ.haifa.ac.il}

\cortext[1]{Corresponding author}

\affiliation[1]{
organization={The Hatter Department of Marine Technologies,
Leon H. Charney School of Marine Sciences,
University of Haifa},
city={Haifa},
postcode={3498838},
country={Israel}
}

%%%%%%%%%%%%%%%%%%%%%%%%%%%%%%%%%%%%%%%%%%%%%%%%%%%%%%%%%%%%%%%%%
%%%%%%%%%%%%%%%%%%%%%%%%%%%%%%%%%%%%%%%%%%%%%%%%%%%%%%%%%%%%%%%%%
%%%%%%%%%%%%%%%%%%%%%%%%%%%%%%%%%%%%%%%%%%%%%%%%%%%%%%%%%%%%%%%%%

\begin{abstract}
Autonomous Underwater Vehicles (AUVs) rely heavily on the fusion of inertial sensors and Doppler velocity logs (DVLs) for navigation. In standard autonomous navigation systems, the DVL measures four beam velocities, thereby enabling the estimation of the AUV velocity vector.
 However, during real-world missions, the DVL may receive noisy or incomplete beam measurements due to marine obstacles, seabed reflections, or environmental disturbances. 
 Furthermore, some low-cost underwater platforms operate without inertial sensors to reduce system complexity and cost. 
 In such cases, reliable estimation of the AUV velocity vector in real-world missing beam scenarios, becomes challenging, leading to degraded navigation solutions.
To circumvent these challenges and enable resilient underwater navigation, we propose DVL-DeepONet, a physics-guided deep neural operator framework along with three variants.
 The proposed models are designed to estimate DVL-based velocity information under multiple operational scenarios, including (i) noise-resilient estimation in coupled inertial/DVL measurements, (ii) DVL-only learning, and (iii) beam measurement recovery. 
By learning a nonlinear operator that maps temporal inertial/DVL observations directly to vehicle velocity while enforcing DVL measurement physics through a consistency constraint, the proposed approach enables robust velocity estimation even under degraded sensing conditions.
The proposed framework is validated using real-world AUV experiments, comprising a cumulative path length of approximately 10,000~m.
 Experimental results demonstrate that the proposed DVL-DeepONet architectures outperform baseline model-based approaches and learning-based algorithms by 40\%.
\end{abstract}


% \begin{highlights}

% \item A physics-guided DVL-DeepONet framework is developed for estimation of DVL velocity vector.

% \item Three DVL-DeepONet variants address noisy, DVL-only, and partial-beam scenarios. 

% \item The framework incorporates the DVL observation model as a physics consistency constraint.

% \item DVL-DeepONet achieves an average improvement of about 40\% over baseline methods.

% \end{highlights}

%\nocite{*}

% Keywords

\maketitle

\section{Introduction}

Autonomous underwater vehicles (AUVs) are increasingly employed in marine missions such as oceanographic surveying, underwater infrastructure inspection, environmental monitoring, and military operations~\cite{paull2013auv,zhang2022submarine}. The success and operational effectiveness of these missions inherently depend on accurate navigation of AUVs. However, this objective remains a significant challenge in deep underwater environments.
%In order to operate autonomously in deep underwater environments, reliable navigation capabilities are essential. 
Since global navigation satellite system signals cannot penetrate underwater, AUVs commonly rely on the fusion of inertial navigation systems (INS) and Doppler velocity logs (DVL) for navigation. The INS provides continuous navigation states using inertial measurements obtained from accelerometers and gyroscopes~\cite{groves2013book,titterton2004strapdown}, while the DVL provides velocity measurements via acoustic Doppler sensing~\cite{wadoo2017autonomous,braginsky2020correction}.

Though INS sensors provide high-rate navigation estimates, their errors accumulate over time due to sensor noise and bias, leading to significant navigation drift~\cite{farrell2007gnss,sahoo2026pidr}.
 To mitigate this, DVL measurements are commonly integrated into the navigation framework as aiding velocity updates. In standard bottom-lock operation, the DVL transmits four acoustic beams toward the seafloor and estimates the vehicle velocity vector by exploiting the Doppler effect. Due to its high velocity accuracy, the DVL plays a critical role in underwater navigation systems~\cite{zhang2023autonomous}.

Over the years, a wide range of velocity estimation strategies have been investigated, including classical model-based approaches such as least-squares (LS) estimator, Kalman filtering, loosely coupled and tightly coupled INS/DVL integration, factor-graph optimization, and information-aided estimation methods~\cite{wang2019novel,engelsman2023information,cheng2026robust}.
Despite their success, robust velocity estimation across diverse underwater operating conditions remains an open challenge. 
In practice, both inertial and DVL measurements are affected by sensor noise, environmental disturbances, and measurement uncertainties, which can significantly degrade estimation accuracy. Furthermore, in certain applications, only DVL measurements may be available, necessitating velocity estimation without inertial data. The problem becomes even more challenging when DVL beam measurements are partially unavailable due to acoustic interference, poor bottom-lock conditions, marine obstacles, rough seafloor terrain, temporary sensor degradation, or aggressive AUV maneuvers~\cite{zhang2026novel}.
 In such circumstances, conventional model-based approaches may experience significant performance degradation, leading to navigation solution drift. 
 
 %Moreover, in traditional model-based navigation, the velocity vector is typically obtained by solving the inverse problem using the measured beam velocities. However, when the beam geometry matrix becomes singular or rank-deficient due to the loss of independent beam measurements, the inverse problem becomes underdetermined. Consequently, the reconstruction becomes ill-conditioned or may fail entirely.

 %Despite these advances, existing methods are often tailored to specific sensing configurations and may struggle to generalize across noisy inertial/DVL measurements, DVL-only operation, and severe beam-loss scenarios within a unified framework.

%%%%%%%%%%%
 To overcome these limitations, recent advances in sensing technologies and computational capabilities have motivated the adoption of data-driven approaches based on machine learning and deep learning. % for underwater navigation and velocity estimation. 
 
 %These methods have demonstrated improved robustness to sensor imperfections, environmental disturbances, and measurement uncertainties compared with traditional model-based techniques.

For inertial/DVL-integrated navigation, several learning-based approaches have been investigated as either end-to-end estimators or hybrid frameworks combined with model-based filtering algorithms. BeamsNet~\cite{cohen2022beamsnet} introduced a neural-network-based velocity estimator that directly maps DVL beam measurements and inertial observations to vehicle velocity. More recently, ResAlignNet~\cite{DAMARI2026125277}, DMIAN~\cite{batovs2026dmian}, and several other hybrid learning-filtering frameworks~\cite{kang2026hybrid,mo2026hybrid} have been explored for inertial/DVL fusion through error state estimation. These methods have demonstrated improved robustness and accuracy over model-based techniques.

For DVL-only navigation tasks, data-driven models have been developed to extract navigation information directly from beam measurements without relying on auxiliary inertial sensors. DCNet~\cite{yampolsky2025dcnet} demonstrated the effectiveness of convolutional neural networks for DVL calibration, while UDON~\cite{zhang2025underwater} explored learning-based underwater odometry using DVL observations. Nevertheless, learning-based approaches for direct DVL-only velocity estimation remain comparatively limited.

Furthermore, under degraded DVL operating conditions, including partial beam availability and beam outages, several learning-based methods have focused on recovering missing information. LiBeamsNet~\cite{cohen2022libeamsnet} addressed limited-beam scenarios by reconstructing missing beam measurements using available beams and inertial data. MissBeamNet~\cite{yona2024missbeamnet} further investigated learning-based velocity estimation under beam outages, while LAS~\cite{zhang2026novel} proposed adaptive strategies for maintaining navigation performance in degraded DVL conditions. More recently, Miao et al.~\cite{miao2026physics} developed a physics-guided long short-term memory (LSTM)-based framework for robust AUV navigation under degraded underwater observations. 
%However, the framework relies on multiple heterogeneous sensors, including INS, DVL, and a polarization compass, which may increase system complexity and deployment cost.

Despite these advances, most existing learning-based approaches are tailored to specific navigation scenarios~\cite{mu2019end,cohen2024seamless}. As a result, a unified learning framework capable of operating across diverse conditions within a single paradigm is still lacking. Furthermore, while these methods have demonstrated promising performance, they generally operate as black-box predictors and do not explicitly exploit the underlying DVL observation model that governs the relationship between beam measurements and vehicle velocity. Consequently, achieving physically consistent and robust velocity estimation across a wide range of operating scenarios, while reducing reliance on expensive external sensors and complex multi-sensor integration architectures, remains an open challenge.

Meanwhile, operator-learning methods have emerged as a powerful paradigm for learning mappings between functional inputs and outputs. Deep Operator Networks (DeepONets), introduced by Lu \textit{et al.} \cite{DeepONet}, have achieved remarkable success in scientific machine learning, surrogate modeling, and dynamical system prediction~\cite{chakraverty2025artificial}. Unlike conventional neural networks, DeepONets preserve the physical properties of the system and offer white-box prediction.
Additionally, it learns operators rather than pointwise mappings, enabling efficient modeling of complex temporal and spatial dependencies. Although DeepONets have been successfully applied to a variety of engineering problems alongside land vehicle dynamics~\cite{tan2025modeling}, their application to underwater navigation and DVL-based velocity estimation remains largely unexplored.

Motivated by these challenges, this research introduces DVL-DeepONet, a physics-guided operator learning framework for underwater velocity estimation.  Unlike conventional DVL navigation methods that rely on explicit matrix inversion, the proposed framework incorporates the DVL observation model solely as a physics-based consistency constraint during training.
 By avoiding direct inversion of the beam geometry matrix, the framework eliminates the rank requirements associated with LS reconstruction. Consequently, DVL-DeepONet enables robust and accurate velocity estimation under severe beam outages and degraded sensing conditions.

Furthermore, the framework is designed to operate under three different sensing conditions. DVL-DeepONet-I addresses noise-resilient estimation using inertial and DVL beam data. DVL-DeepONet-II focuses on DVL only beam-to-velocity operator learning and enables velocity estimation even in the absence of inertial measurements. 
DeepONet-III handles scenarios involving partial or missing beam configurations.
By integrating (a) physical consistency, (b) temporal sensor fusion, and (c) operator learning, the proposed approach aims to provide accurate and resilient velocity estimation for real-world underwater navigation.

 The contributions of this research are outlined as follows:

\begin{enumerate}

\item A novel physics-guided DVL-DeepONet framework, uniquely designed to forecast DVL velocity vector.

%that incorporates an LS consistency constraint during training.

\item Three complementary DVL-DeepONet architectures to address practical underwater navigation scenarios.

\item Extensive validation on AUV datasets collected from real-world sea trials and competitive baselines, demonstrating improved robustness, velocity estimation accuracy, and navigation performance. 

\item To support the reproduction of results and encourage future research and benchmarking, our codebase has been made publicly available on \url{https://github.com/ansfl/DVL-DeepONet}.

\end{enumerate}

The proposed framework is validated using real-world AUV datasets collected during sea trials in the Mediterranean Sea, Israel, comprising approximately 10,000~m of underwater navigation trajectories.
 Extensive experiments are conducted under noisy measurements, partial beam outages, and limited sensing scenarios. The results demonstrate that the proposed DVL-DeepONet framework provides accurate and resilient velocity estimation under challenging underwater conditions. Our approach offers an average improvement of 40\% over baselines.

The remainder of this paper is organized as follows. Section~\ref{sec:2} presents the problem formulation and DVL measurement model. Section~\ref{sec:3} introduces the DVL-DeepONet framework and network architectures.  Section~\ref{sec:4} presents the experimental results and analysis. Finally, Section~\ref{sec:6} concludes the paper.

%%%%%%%%%%%%%%%%%%%%%%%%%%%%%%%%%%%%%%%%%%%%%%%%%%%%%%

% =========================================================
\section{Problem Formulation} \label{sec:2}

 This section formulates the DVL velocity estimation problem and highlights the challenges associated with limited and degraded beam measurements.

\subsection{Least-Squares Velocity Reconstruction}
\begin{figure*}[t] 
    \begin{center}
        \includegraphics[height=8cm,width=10cm]{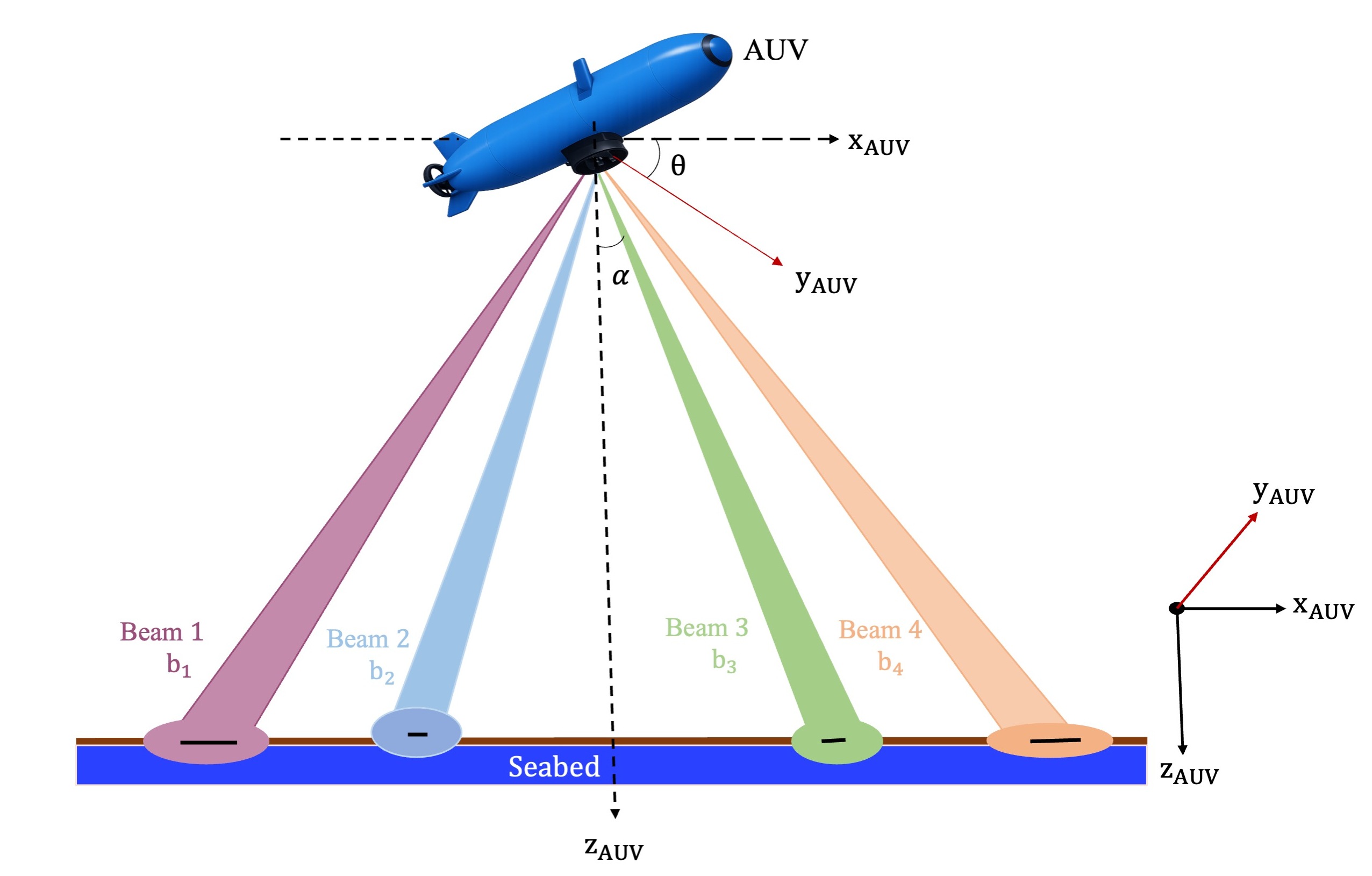}
        \caption{Geometry of the four-beam Janus DVL mounted on the AUV and the associated body-fixed coordinate system.}
         \label{fig:1}
    \end{center} 
\end{figure*}

A standard DVL employs four acoustic transducers commonly arranged in a Janus (X) configuration. Each transducer emits an acoustic beam toward the seafloor as depicted in Fig.~\ref{fig:1} and measures the Doppler frequency shift of the reflected signal~\cite{brokloff1994matrix}. Let $\mathbf b_i,\; i=1,2,3,4,$ denote the beam direction vectors defined in the DVL frame.
 Thus, the DVL projection matrix is obtained as

\begin{equation} \label{eq:16}
\mathbf H
=
\begin{bmatrix}
\mathbf b_1\\
\mathbf b_2\\
\mathbf b_3\\
\mathbf b_4
\end{bmatrix}
\in\mathbb R^{4\times3}.
\end{equation}

%%%%%%%%%%%%%%%%%%%%%%%%%%%%%%%%%%%%%%%%%%%

%%%%%%%%%%%%%%%%%%%%%%%%%%%%%%%%%%%%%%%%%%%

\noindent
% Let the AUV velocity vector expressed in the DVL frame is 

% \begin{equation}
% \mathbf v_b^d
% =
% \begin{bmatrix}
% v_x & v_y & v_z
% \end{bmatrix}^{T} \in\mathbb R^{3\times1}.
% \end{equation}
\noindent The beam measurement model is~\cite{liu2018ins}

\begin{equation} \label{eq:5}
\mathbf y
=
\mathbf H\mathbf v_b^d
+
\mathbf n,
\end{equation}
where  $\mathbf n$ denotes measurement noise, $\mathbf y$ is the measured beam velocity, and $\mathbf v_b^d$ is the AUV velocity vector expressed in the DVL coordinate frame.

% \begin{equation}
% \mathbf y
% =
% \begin{bmatrix}
% y_1 & y_2 & y_3 & y_4
% \end{bmatrix}^{T}
% \end{equation}
% contains the measured beam velocities 

The objective of the DVL processing stage is to reconstruct the velocity vector from the beam measurements. 
The standard DVL solution employs a LS estimator to solve \eqref{eq:5}. The estimation problem can be formulated as~\cite{braginsky2020correction}:

\begin{equation} \label{eq:0}
\hat{\mathbf v}_b^d
=
\arg\min_{\mathbf v}
\left\|
\tilde{\mathbf y}-\mathbf H\mathbf v_b^d
\right\|_2^2.
\end{equation}

\noindent The solution to \eqref{eq:0} is
% The corresponding normal equation is

% \begin{equation}
% \mathbf H^{T}\mathbf H\hat{\mathbf v}_b^d
% =
% \mathbf H^{T}\tilde{\mathbf y}.
% \end{equation}
% \noindent
% Assuming $\mathbf H^{T}\mathbf H$ is nonsingular, the solution becomes

\begin{equation} \label{eq:1}
\hat{\mathbf v}_b^d
=
(\mathbf H^{T}\mathbf H)^{-1}
\mathbf H^{T}
\tilde{\mathbf y}.
\end{equation}
\noindent
% Define the pseudoinverse matrix

% \begin{equation}
% \mathbf H^{\dagger}
% =
% (\mathbf H^{T}\mathbf H)^{-1}
% \mathbf H^{T}.
% \end{equation}
% \noindent
% Now, the velocity estimate \eqref{eq:1} can be  compactly written as

% \begin{equation}
% \hat{\mathbf v}_b^d
% =
% \mathbf H^{\dagger}\tilde{\mathbf y}.
% \end{equation}

% The residual error of the LS solution is

% \begin{equation}
% \mathbf r
% =
% \tilde{\mathbf y}
% -
% \mathbf H\hat{\mathbf v}_b^d,
% \end{equation}

% with associated cost function

% \begin{equation}
% J(\hat{\mathbf v}_b^d)
% =
% \mathbf r^{T}\mathbf r.
% \end{equation}

Although computationally efficient, the LS solution is sensitive to noisy beam measurements, outliers, and limited beam availability. In practical underwater environments, these issues may significantly degrade the algorithm performance.

\subsection{Navigation Under Limited DVL Measurements}

In a DVL-missing-beams, a beam availability vector at time $k$ is defined as

\begin{equation} \label{eq:10}
\mathbf m_k
=
[m_{1,k},m_{2,k},m_{3,k},m_{4,k}]^{T},
\qquad
m_{i,k}\in\{0,1\},
\end{equation}

% where

% \begin{equation}
% m_{i,k}
% =
% \begin{cases}
% 1, & \text{beam available},\\
% 0, & \text{beam unavailable}.
% \end{cases}
% \end{equation}
\noindent
The corresponding masking matrix is

\begin{equation}  \label{eq:11}
\mathbf M_k
=
\mathrm{diag}
(\mathbf m_k) \in
\mathbb R^{4\times4}.
\end{equation}

\noindent
The partially observed beam measurements are

\begin{equation} \label{eq:3}
\mathbf y_k^p
=
\mathbf M_k \mathbf y_k.
\end{equation}

\noindent
Substituting the beam measurement model \eqref{eq:5} into the \eqref{eq:3} yields

\begin{equation}
\mathbf y_k^p
=
\mathbf M_k
\left(
\mathbf H \mathbf v_{b,k}^{d}
+
\mathbf n_k
\right).
\end{equation}

\noindent
The number of available beams is

\begin{equation}
N_{b,k}
=
\sum_{i=1}^{4}m_{i,k}.
\end{equation}
\noindent
When $N_{b,k} \ge 3$
a reduced LS solution may still be computed. However, when
$N_{b,k} < 3,$
the velocity estimation problem becomes underdetermined, as

\begin{equation}
\mathrm{rank}
(\mathbf H_k^p)
<
3.
\end{equation}
Here $\mathbf H_k^p\in\mathbb R^{N_{b,k}\times3}$
denotes the reduced beam projection matrix obtained by
removing the rows of $\mathbf H$ corresponding to unavailable beams at time step \(k\).

Consequently, a unique velocity solution no longer exists, and conventional LS estimation becomes unreliable or infeasible. These challenges motivate the development of robust approaches capable of exploiting both inertial measurements and partial beam observations to estimate the AUV velocity under degraded sensing conditions.

\section{Proposed DVL-DeepONet Framework} \label{sec:3}
 This section discusses  a family of physics-guided operator learning models, termed DVL-DeepONet, for resilient underwater velocity estimation under varying sensor availability conditions.

DeepONet was selected as the backbone architecture because the underwater navigation problem considered in this work can be naturally formulated as an operator-learning task.
DeepONet is a neural operator framework that is even capable of learning mappings between infinite-dimensional function spaces. Its formulation is mathematically grounded in the universal approximation theorem for operators~\cite{DeepONet}. 
Rather than estimating velocity from a single sensor snapshot, the objective is to learn a mapping from a temporal history of beam measurements to the corresponding AUV velocity vector.

To further improve physical consistency, DVL-DeepONet incorporates an LS beam consistency constraint during training. It ensures that the predicted velocity remains compatible with the underlying DVL measurement geometry.
We begin by presenting DVL-DeepONet and then address three different real-world scenarios: (i) noise-resilient estimation, (ii) DVL-only learning, and (iii) beam measurement recovery.

 % This combination of operator learning and physics-guided regularization is particularly attractive for underwater navigation problems involving noisy sensors and partial beam availability,

%While DeepONet was originally introduced for function-to-function operator learning, the proposed navigation problem may be interpreted as a finite-dimensional operator-learning task in which a temporal history of sensor measurements is mapped to the corresponding vehicle velocity. This perspective motivates the use of DeepONet as the backbone architecture of the proposed framework.
 
%In particular, DVL-DeepONet-II achieves the lowest RMSE under partial DVL beam availability, indicating that the proposed branch--trunk operator-learning formulation is more effective in capturing the nonlinear relationship between incomplete beam measurements, inertial observations, and vehicle velocity.

\subsection{DVL-DeepONet Framework}

 The proposed framework integrates DeepONet with DVL beam geometry and LS physical constraints to estimate the velocity vector under noisy, degraded, and partially observable sensing conditions.

Let $\mathbf y_k$ denote the DVL beam measurements at time step $k$,

\begin{equation}
\mathbf y_k
=
[y_{1,k},y_{2,k},y_{3,k},y_{4,k}]^T
\in
\mathbb R^{4},
\end{equation}
where $y_{i,k}$ is the velocity measured by the $i$-th DVL beam.
Similarly, let $\mathbf u_k$ denote the synchronized IMU measurement vector,

\begin{equation}
\mathbf u_k
=
[f_{x,k},f_{y,k},f_{z,k},
\omega_{x,k},\omega_{y,k},\omega_{z,k}]^T
\in
\mathbb R^{6},
\end{equation}
where $f_{x,k}$, $f_{y,k}$, and $f_{z,k}$ are the averaged specific-force components, and $\omega_{x,k}$, $\omega_{y,k}$, and $\omega_{z,k}$ are the averaged angular-rate components associated with the $k$-th DVL interval.
To align with the DeepONet structure, the inertial readings are averaged between two successive DVL measurements.

To capture temporal dependencies, a sliding window of length $W$ is employed. The corresponding DVL and IMU histories are

\begin{equation}
\mathbf Y_k
=
[\mathbf y_{k-W+1},
\ldots,
\mathbf y_k]
\in
\mathbb R^{W\times4},
\end{equation}

\begin{equation}
\mathbf U_k
=
[\mathbf u_{k-W+1},
\ldots,
\mathbf u_k]
\in
\mathbb R^{W\times6},
\end{equation}
where $k$ denotes the current discrete-time index.

DVL-DeepONet architecture consists of branch and trunk networks. The branch input $\mathbf B_k$ is constructed by channel-wise concatenation,

\begin{equation} \label{eq:9}
\mathbf B_k
=
[
\mathbf Y_k \, || \, \mathbf U_k
]
\in
\mathbb R^{W\times D},
\end{equation}
where $||$ denotes concatenation and $D$ denotes the number of synchronized sensor channels. For the considered DVL--IMU configuration, $D = 4 + 6 = 10$ channels.

The synchronized sensor channels are jointly processed by the branch encoder to learn a latent representation of the recent vehicle dynamics.
 The objective is to learn a nonlinear operator

\begin{equation}
\mathcal G_\theta:
(\mathbf B_k,\tau_k)
\rightarrow
\mathbf v_k,
\end{equation}
where $\mathbf v_k $ denotes the velocity vector:

\begin{equation}
\mathbf v_k
=
[v_x,v_y,v_z]^T
\in
\mathbb R^3,
\end{equation}
\noindent
and $\tau_k$ is the normalized temporal coordinate defined by:

\begin{equation}
\tau_k
=
\frac{t_k-\mu_t}{\sigma_t}.
\end{equation}

\noindent
The branch network $\mathbf B_k$ encodes the synchronized IMU/DVL history into a latent representation.
$\mathbf B_k$ is implemented as a one-dimensional (1-D) convolutional encoder. Since the input to the convolutional layers is arranged as channels by time, the branch input is internally converted to

\begin{equation}
\widetilde{\mathbf B}_k
=
\mathbf B_k^{\top}
\in
\mathbb R^{D\times W}.
\end{equation}

\noindent Then, the convolutional encoder is defined as

\begin{equation}
\mathbf h_b^{(l)}
=
\sigma
\left(
\mathbf W_b^{(l)}
*
\mathbf h_b^{(l-1)}
+
\mathbf b_b^{(l)}
\right),
\qquad
l=1,2,3,
\end{equation}
\noindent

\noindent
where \(
\mathbf h_b^{(0)}=\widetilde{\mathbf B}_k
\)
denotes the branch input,
\(
\mathbf h_b^{(l)}
\)
is the feature map of the \(l\)-th convolutional layer,
\(
\mathbf W_b^{(l)}
\)
and
\(
\mathbf b_b^{(l)}
\)
represents the trainable convolution kernels and bias parameters, respectively, 
\(( *) \) denotes 1-D convolution along with the temporal axis,
and
\(
\sigma(\cdot)
\)
is the nonlinear activation function.

The output of the final convolutional layer is flattened and passed through a fully connected projection layer to obtain the branch latent representation
\begin{equation}
\mathbf z_b
=
\phi_b
\left(
\mathrm{vec}
\left(
\mathbf h_b^{(3)}
\right)
\right)
\in
\mathbb R^{p},
\end{equation}
where
\(
\mathrm{vec}(\cdot)
\)
denotes the flattening operation,
\(
\phi_b(\cdot)
\)
represents the fully connected projection, and $p$ denotes the latent embedding dimension of the DeepONet architecture.

Let \(\mathcal B_{\theta_b}\) denote the branch encoder with trainable parameters \(\theta_b\).
It consists of three 1-D convolutional layers with 32, 64, and 64 output channels, respectively.
  Equivalently, the branch encoder can be written compactly as

\begin{equation}
\mathbf z_b
=
\mathcal B_{\theta_b}
(\mathbf B_k)
\in
\mathbb R^{p}.
\end{equation}

\noindent
Thus, the branch network explicitly fuses DVL beam measurements and IMU observations through channel-wise concatenation before temporal convolutional feature extraction.

%%%%%%%%%%%%%%

Next, the trunk network maps the normalized temporal coordinate
\(
\tau_k
\)
into the latent operator space through a multilayer perceptron defined as

\begin{equation}
\mathbf h_t^{(l)}
=
\phi
\left(
\mathbf W_t^{(l)}
\mathbf h_t^{(l-1)}
+
\mathbf b_t^{(l)}
\right),
\qquad l=1,\ldots,L,
\end{equation}

\noindent
where
\(
\mathbf h_t^{(0)}=\tau_k
\),
\(
\mathbf W_t^{(l)}
\)
and
\(
\mathbf b_t^{(l)}
\)
denote the trainable weight matrices and bias vectors, respectively, and
\(
\phi(\cdot)
\)
is the nonlinear activation function.
The output of the final hidden layer is projected to the trunk latent representation

\begin{equation}
\mathbf z_t
=
\mathcal T_{\theta_t}
(\tau_k)
\in
\mathbb R^{p},
\end{equation}
where \(\mathcal T_{\theta_t}\) is the trunk encoder MLP parameterized by \(\theta_t\).

 In the implemented architecture, the trunk encoder maps the scalar temporal input to a latent representation of dimension \(p=128\).
The trunk encoder consists of three fully connected layers with
64, 128, and 128 neurons, respectively, with tanh activation function.
Following the DeepONet principle, the branch and trunk embeddings are fused through element-wise multiplication

\begin{equation}
\mathbf z_k
=
\mathbf z_b
\odot
\mathbf z_t .
\end{equation}
%%%%%%%%%%%%%%%%%%%%%%%

\noindent
Furthermore, the fused latent representation is subsequently processed by a prediction head MLP consisting of two hidden fully connected layers with 128 and 64 neurons, respectively, followed by an output layer with three neurons corresponding to the velocity components. The prediction head employs the sigmoid linear unit (SiLU) activation function in the hidden layers.
It may be compactly written as
%\noindent The fused representation is subsequently mapped to Cartesian velocity space,

\begin{equation}
\hat{\mathbf v}_k
=
\mathcal H_{\theta_h}
(\mathbf z_k),
\end{equation}
%where \(\mathcal H_{\theta_h}(\cdot)\) maps the latent operator representation to the Cartesian velocity space.
where
\(
\mathcal H_{\theta_h}(\cdot)
\)
denotes the prediction head MLP parameterized by
\(
\theta_h
\), and $\hat{\mathbf v}_k$ is the predicted vehicle velocity vector.
 Therefore, the DVL-DeepONet operator approximation results in

\begin{equation}
\hat{\mathbf v}_k
=
\mathcal H_{\theta_h}
\left(
\mathcal B_{\theta_b}(\mathbf B_k)
\odot
\mathcal T_{\theta_t}(\tau_k)
\right).
\end{equation}
%\noindent This fusion mechanism follows the idea of DeepONet, coupling a branch representation of the input function with a trunk representation of the evaluation coordinate. 
%The fused latent representation is subsequently processed by a fully connected prediction head, which maps the operator embedding into the three-dimensional Cartesian velocity space.
%%%%%%%%%%%%%%%%%%%%%%%%%%

\begin{figure*}[!ht]
    \centering
    \includegraphics[width=0.8\textwidth]{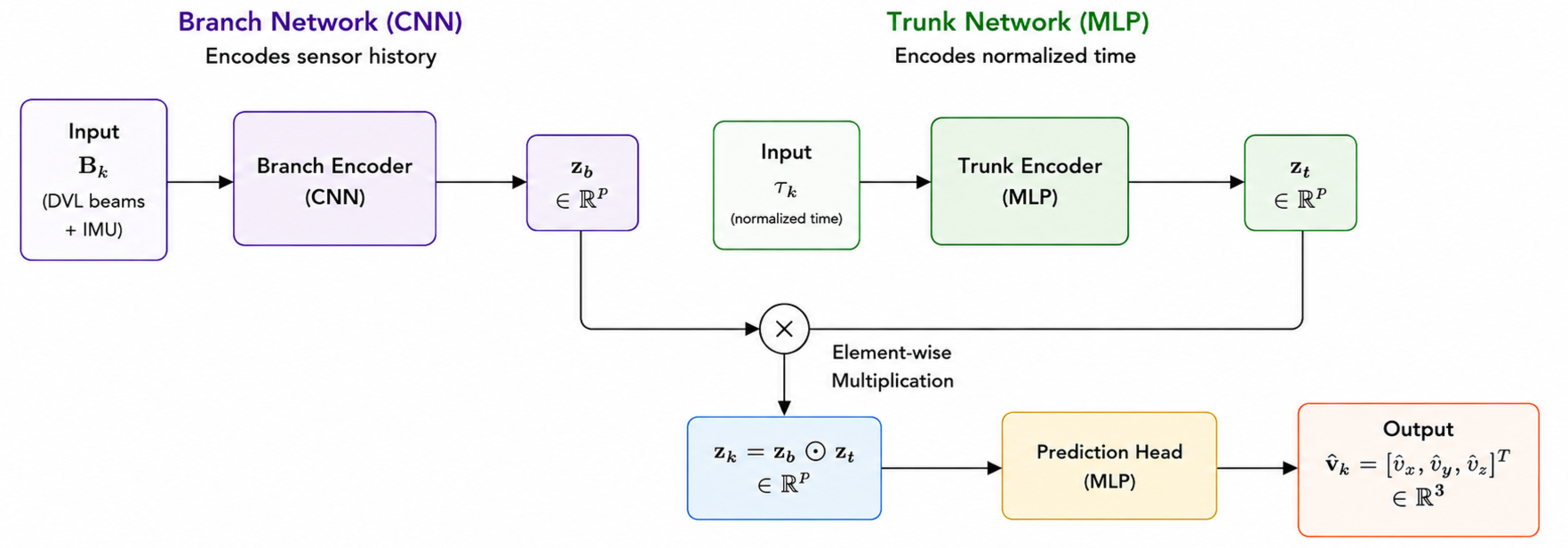}
    \caption{Schematic illustration of the proposed DVL-DeepONet architecture.}
    \label{fig:ls_deeponet}
\end{figure*}

The proposed framework consists of three complementary operator-learning architectures, which are designed o handle noisy DVL and inertial measurements (DVL-DeepONet-I), DVL-only operation (DVL-DeepONet-II), and partial sensor availability (DVL-DeepONet-III), respectively, as shown in Table~\ref{tab:lsdeeponet_variants}. The overall framework is illustrated in Fig.~\ref{fig:ls_deeponet}. 
In all three DVL-DeepONet variants, the framework exploits temporal sensor histories and nonlinear operator learning to capture the underlying beam-to-velocity dynamics. 
%%%%%%%%%%%%%%%%%%%%%%%%%%%%%%%%%%%%%%%%%%%%%%%%%%%
\begin{table}[t]
\centering
\caption{Summary of DVL-DeepONet variants for estimating the DVL velocity vector in different conditions.}
\label{tab:lsdeeponet_variants}
\begin{tabular}{lccc}
\toprule
Model & IMU & DVL & Objective \\
\midrule
DVL-DeepONet-I & \checkmark & \checkmark &
Noise-Resilient \\

DVL-DeepONet-II & $\times$ & \checkmark &
DVL-Only \\

DVL-DeepONet-III & \checkmark & Partial &
Beam Recovery \\
\bottomrule
\end{tabular}
\end{table}

%%%%%%%%%%%%%%%%%%%%%%%%%%%%%%%%%%%%%%%%%%%%%%%%%%%
\noindent
%Unlike conventional least-squares reconstruction, which independently estimates velocity at each measurement epoch, t

% A least-squares beam-consistency constraint and a velocity-domain Kalman filter are incorporated to improve physical consistency and temporal smoothness. 

% ============================================================
\subsection{Noise-Resilient Estimation}

DVL-DeepONet-I is designed for robust velocity estimation under noisy IMU and DVL measurements. 
%The branch input consists of synchronized IMU and DVL histories.
For a temporal window of length \(W\), the branch input \eqref{eq:9} is constructed as

\begin{equation}
\mathbf B_k
=
[
\mathbf y_{k-W+1},\ldots,\mathbf y_k,
\mathbf u_{k-W+1},\ldots,\mathbf u_k
]
\in
\mathbb R^{W\times 10}.
\end{equation}

\noindent The network learns the nonlinear operator

\begin{equation}
\mathcal G_{\theta}^{(I)}
:
(\mathbf B_k,\tau_k)
\rightarrow
\mathbf v_k.
\end{equation}

% By jointly exploiting inertial measurements and DVL observations, DVL-DeepONet-I provides robust velocity estimation under measurement noise and sensor uncertainties.

% \subsubsection{Training Strategy}

% \noindent
%  DVL-DeepONet-I variants are trained using the same supervised operator-learning framework.
\noindent For a mini-batch of size $N$, the supervised loss is defined as

\begin{equation} \label{eq:6}
\mathcal L_{vel}
=
\frac{1}{N}
\sum_{i=1}^{N}
\|
\hat{\mathbf v}_i
-
\mathbf v_i
\|_2^2.
\end{equation}
\noindent
To enforce consistency with the DVL observation model introduced in \eqref{eq:5}, the predicted velocity is projected back into beam space,

\begin{equation}
\hat{\mathbf y}_i
=
\mathbf H
\hat{\mathbf v}_i.
\end{equation}
where \(\mathbf H\in\mathbb R^{4\times3}\) denotes the DVL beam projection matrix \eqref{eq:16}.
% The corresponding least-squares beam consistency loss is
% \begin{equation}
% \mathcal L_{LS}
% =
% \frac{1}{N}
% \sum_{i=1}^{N}
% \left\|
% \mathbf H
% \hat{\mathbf v}_i
% -
% \mathbf y_i
% \right\|_2^2.
% \end{equation}
Building on the DVL beam projection matrix, the residual is defined as

\begin{equation}
\mathbf r_i
=
\mathbf y_i
-
\mathbf H
\hat{\mathbf v}_i,
\end{equation}
\noindent
Thereby, the physics-informed loss can be written as

\begin{equation} \label{eq:7}
\mathcal L_{LS}
=
\frac{1}{N}
\sum_{i=1}^{N}
\mathbf r_i^{T}
\mathbf r_i.
\end{equation}

\noindent Combining \eqref{eq:6} and \eqref{eq:7} leads to the total loss

\begin{equation} \label{eq:8}
\mathcal L
=
\lambda_{vel}
\mathcal L_{vel}
+
\lambda_{LS}
\mathcal L_{LS}.
\end{equation}
where \(\lambda_{vel}\) and \(\lambda_{LS}\) denote positive weighting coefficients. Based on extensive hyperparameter tuning through a series of trial-and-error experiments, the coefficients were fixed at
 \(\lambda_{vel}=1\) and \(\lambda_{LS}=0.1\).

The LS consistency term acts as a geometry-aware physical regularizer that constrains the learned velocity to remain compatible with the DVL beam observation model.
The optimized network parameters satisfy 

\begin{equation}
\Theta^{*}
=
\arg\min_{\Theta}
\mathcal L(\Theta).
\end{equation}

% ============================================================
\subsection{DVL-Only Learning}

DVL-DeepONet-II addresses scenarios where IMU measurements are unavailable or unreliable. In this case, the branch input contains only DVL beam observations.
 For a temporal window of length \(W\), the branch input is constructed as

\begin{equation}
\mathbf B_k
=
\left[
\mathbf y_{k-W+1},
\mathbf y_{k-W+2},
\ldots,
\mathbf y_k
\right]
\in
\mathbb R^{W\times4}.
\end{equation}

\noindent
The learned operator becomes

\begin{equation}
\mathcal G_{\theta}^{(II)}
:
(\mathbf B_k,\tau_k)
\rightarrow
\mathbf v_k.
\end{equation}

 % DVL-DeepONet-II exploits temporal beam dynamics to estimate Cartesian velocity directly from DVL measurements without requiring inertial information.

\noindent  DVL-DeepONet-II is trained using the same supervised loss \eqref{eq:6}, and physics-informed loss \eqref{eq:7}.

% ============================================================
\subsection{Beam Measurement Recovery}

DVL-DeepONet-III is developed for navigation under DVL beam outages. A key distinction between the proposed framework and conventional model-based DVL velocity reconstruction lies in the treatment of missing beam measurements. Unlike the traditional LS estimator formulation in (\ref{eq:1}), DVL-DeepONet-III does not estimate velocity through matrix inversion. Instead, the DVL observation model is incorporated as a physics-guided constraint, while the velocity is inferred through operator learning. Consequently, the framework is not restricted by the rank requirements of conventional LS reconstruction and can operate even when only one or two DVL beams are available. The available beam measurement provides partial physical information, whereas the missing information is recovered from synchronized IMU observations, temporal context, and previously learned beam-to-velocity mappings. This enables accurate velocity estimation even under severe beam outages.

 %In practical underwater environments, one or more DVL beams may become unavailable due to acoustic interference, seafloor discontinuities, marine obstacles, sediment clouds, or aggressive AUV maneuvers. 
 %Under such conditions, conventional least-squares reconstruction becomes unreliable or completely fails when the number of available beams is insufficient.

% \begin{equation}
% \mathcal L_{LS}
% =
% \left\|
% \mathbf H\hat{\mathbf v}
% -
% \mathbf y
% \right\|_2^2,
% \label{eq:ls_constraint}
% \end{equation}

% In classical DVL navigation, the Cartesian velocity vector is obtained by solving the inverse problem \eqref{eq:1}

% Consequently, a unique velocity solution requires

% \begin{equation}
% \mathrm{rank}(\mathbf H)\geq 3.
% \end{equation}

% When fewer than three independent beams are available, the inverse problem becomes underdetermined and the LS reconstruction becomes ill-conditioned or fails entirely.

% The partially observed beam measurements are

% \begin{equation}
% \mathbf y_k^p
% =
% \mathbf M_k \mathbf y_k.
% \end{equation}

Using the partial DVL measurements in (\ref{eq:3}) and the synchronized IMU observations in (\ref{eq:9}), the branch input is evolved as 

\begin{equation}
\mathbf B_k
=
[
\mathbf y_{k-W+1}^{p},
\ldots,
\mathbf y_k^{p},
\mathbf u_{k-W+1},
\ldots,
\mathbf u_k
].
\end{equation}

\noindent
The corresponding operator is

\begin{equation}
\mathcal G_{\theta}^{(III)}
:
(\mathbf B_k,\tau_k)
\rightarrow
\mathbf v_k.
\end{equation}

\noindent
The supervised velocity loss associated with partial beams is defined as

\begin{equation} \label{eq:13}
\mathcal L_{p-vel}
=
\frac{1}{N}
\sum_{k=1}^{N}
\|
\hat{\mathbf v}_k
-
\mathbf v_k
\|_2^2.
\end{equation}

\noindent
To enforce consistency with the partial DVL beam measurements, a masked physics-informed loss is introduced as

\begin{equation} \label{eq:14}
\mathcal L_{p-LS}
=
\frac{1}{N}
\sum_{k=1}^{N}
\sum_{i=1}^{4}
m_{i,k}
(\hat y_{k,i}-y_{k,i})^2,
\end{equation}

% where

% \[
% m_{i,k}
% =
% \begin{cases}
% 1, & \text{if beam } i \text{ is available},\\
% 0, & \text{otherwise}.
% \end{cases}
% \]
\noindent
Consequently, the total loss is

\begin{equation} \label{eq:15}
\mathcal L
=
\lambda_{p-vel}\mathcal L_{p-vel}
+
\lambda_{p-LS}\mathcal L_{p-LS},
\end{equation}
where
$\lambda_{p-vel},\lambda_{p-LS}>0$
are weighting coefficients.

%By incorporating synchronized IMU measurements and partial beam observations, DVL-DeepONet-III enables velocity estimation under degraded sensing conditions and DVL beam outages.

% ============================================================

\subsection{Training}
Algorithm~\ref{alg:ls_deeponet} summarizes the complete training workflow. 
The proposed DVL-DeepONet framework was trained using the hyperparameters detailed in Table~\ref {tab:all_configs}.

%%%%%%%%%%%%%%%%%%%%%%%%%%%

\begin{algorithm}[!ht]
\caption{Training procedure of DVL-DeepONet (identical for all its three variants).}
\label{alg:ls_deeponet}

Construct branch input $\mathbf B_k$\;
Construct trunk input $\tau_k$\;
Initialize network parameters $\Theta$\;

\For{each epoch}{
    \For{each mini-batch}{
        Compute $\mathbf z_b=\mathcal B_{\theta_b}(\mathbf B_k)$\;
        Compute $\mathbf z_t=\mathcal T_{\theta_t}(\tau_k)$\;
        Fuse $\mathbf z_k=\mathbf z_b\odot\mathbf z_t$\;
        Predict $\hat{\mathbf v}_k=\mathcal H_{\theta_h}(\mathbf z_k)$\;
        Compute $\mathcal L_{\mathrm{vel}}$ using \eqref{eq:6}, or \eqref{eq:13}  in the missing beam scenario\;
        Compute $\mathcal L_{\mathrm{LS}}$ via \eqref{eq:7} or \eqref{eq:14}  in the missing beam scenario\;
        Compute total loss $\mathcal L$ using \eqref{eq:8} or \eqref{eq:15}  in the missing beam scenario\;
        Update $\Theta$ using backpropagation\;

        \If{$\mathcal L < \varepsilon$}{
            \Return{$\Theta$}
        }
    }
}

\Return{$\Theta$}
\end{algorithm}

%%%%%%%%%%%%%%%%%%%%%%%%%%%

\begin{table}[htbp]
\centering
\caption{Hyperparameters for DVL-DeepONet framework used in all three variants.}
\label{tab:all_configs}
\footnotesize
\begin{tabular}{ll}
\toprule
\textbf{Parameter} & \textbf{Configuration} \\
\midrule

Temporal Window ($W$) & 2 \\

Branch CNN & 32-64-64 \\

Trunk MLP & 64-128-128 \\

Prediction Head MLP& 128-64-3 \\

Optimizer & AdamW \\

Learning Rate & $10^{-3}$ \\

Batch Size & 128 \\

Epochs & 100 \\

\bottomrule
\end{tabular}
\end{table}

% The corresponding least-squares beam consistency loss is

% \begin{equation}
% \mathcal L_{LS}
% =
% \frac{1}{N}
% \sum_{i=1}^{N}
% \left\|
% \mathbf H
% \hat{\mathbf v}_i
% -
% \mathbf y_i
% \right\|_2^2.
% \end{equation}

% The physics-informed loss becomes

% \begin{equation}
% \mathcal L_{LS}
% =
% \frac{1}{N}
% \sum_{i=1}^{N}
% \|
% \mathbf H
% \hat{\mathbf v}_i
% -
% \mathbf y_i
% \|_2^2.
% \end{equation}

% \noindent
% Finally, the predicted velocity sequence is refined through a velocity-domain Kalman filter,

% \begin{equation}
% \hat{\mathbf v}_{KF}
% =
% \mathrm{KF}
% (
% \hat{\mathbf v}
% ),
% \end{equation}
% which suppresses high-frequency estimation noise and improves temporal consistency.

% =========================================================

\section{Analysis and Results} \label{sec:4}
This section discusses the dataset, evaluation metrics, and results obtained by DVL-DeepONet. 
%We have considered different baselines to evaluate the performance of our three DVL-DeepONet variants, as there is no one-size-fits-all model for all three operational conditions of the AUV. The claims are further supported by an ablation study.

\subsection{AUV Dataset }

The AUV dataset was acquired using the University of Haifa's Snapir AUV during multiple sea trials conducted in the Mediterranean Sea, Israel \cite{shurin2022autonomous}. %as shown in Fig.~\ref{fig:auv}. 

% \begin{figure}[!ht]
%     \begin{center}
%         \includegraphics[height=7cm,width=10cm]{auv_new.png}
%         \caption{Snapir AUV being lowered from a ship into the Mediterranean Sea for mission.
% }
%         \label{fig:auv}
%     \end{center}
% \end{figure}

\begin{figure}
    \centering
    \includegraphics[width=0.95\columnwidth]{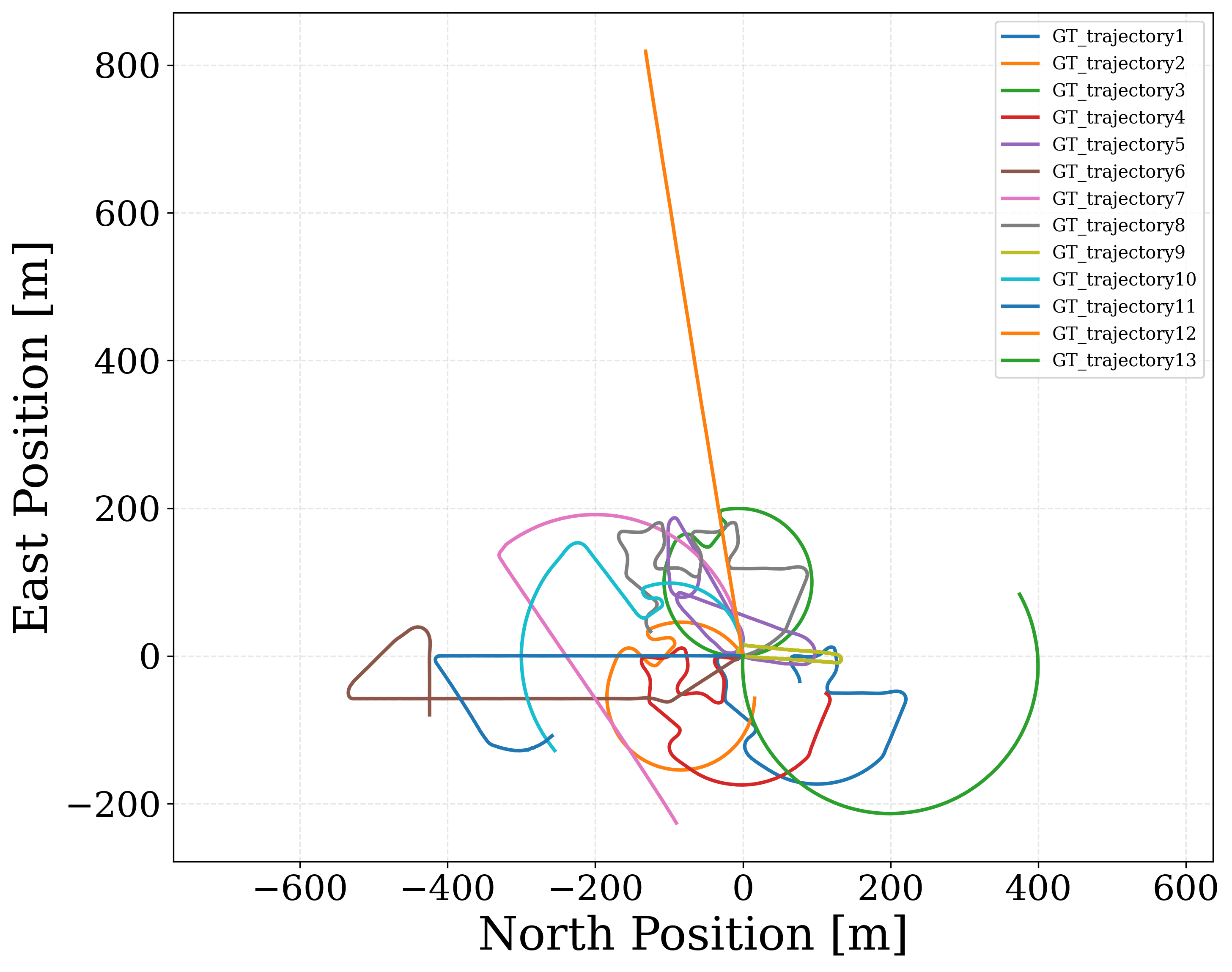}
    \caption{Top-view of the 13 AUV trajectories used in this study.}
    \label{fig:all_traj}
\end{figure}

Snapir is an \cite{eca_a18d_2023} ECA Robotics modified A18D mid-size AUV designed for deep-water operations up to 3000~m and capable of missions lasting up to 21~hours.
The platform IS mounted with an iXblue Phins Subsea INS \cite{ixblue_phins_2023}, and a Teledyne RDI WorkHorse Navigator DVL \cite{teledyne_dvl_2023}, which provides velocity measurements with a nominal standard deviation of 0.02~m/s. The INS operates at 100~Hz, whereas the DVL provides measurements at 1~Hz.
The dataset consists of multiple AUV maneuvers, each spanning approximately 400~s. These missions differ in path geometry, mission lengths, operating depth, and vehicle speed, as presented in Fig.~\ref{fig:all_traj}. 
These thirteen trajectories provide a comprehensive evaluation of the proposed DVL-DeepONet framework under varying operating conditions.

For cross-validation purposes, we partition the dataset into three distinct splits. In the first, the training set comprises T1,  T6--T13, and the validation set comprises T2 and T3. The remaining two trajectories, T4 and T5, are reserved for testing. The total path lengths of the training and testing datasets are approximately 7095~m and 1566~m, respectively.
To maintain clarity and brevity, only the results and graphs obtained from the primary experimental configuration (Split 1) are reported in detail. More information regarding three-fold cross-validation splits are discussed later in Section \ref{sec:cross}.

To construct the noisy DVL measurements, the DVL velocity measurements recorded during the AUV missions were first projected onto the four DVL beam directions using the beam geometry matrix. The resulting beam measurements are given by

\begin{equation}
\mathbf y_k
=
\mathbf H \mathbf v_{b,k}^{d},
\end{equation}
where \(\mathbf H\in\mathbb R^{4\times3}\) denotes the DVL beam geometry matrix and \(\mathbf v_{b,k}^{d}\) is the DVL velocity vector expressed in the body frame.

To emulate practical underwater environments, scale-factor error, constant bias, and additive Gaussian noise were introduced into the beam measurements 

\begin{equation}
\tilde{\mathbf y}_k
=
(1+s_{DVL})
\mathbf H
\mathbf v_{b,k}^{d}
+
\mathbf b_{DVL}
+
\sigma_{DVL}\boldsymbol{\epsilon}_k,
\end{equation}
where \(s_{DVL}=0.7\%\) is the scale-factor error, \(\mathbf b_{DVL}=0.001\,\mathbf 1_4\) is the bias vector, \(\sigma_{DVL}=0.042\) is the noise standard deviation, and $\boldsymbol{\epsilon}_k
\sim
\mathcal N(\mathbf 0,\mathbf I_4)$.
The specific error term values follow those used in~\cite{cohen2022beamsnet}.
To this end, the corrupted beam measurements \(\tilde{\mathbf y}_k\) were subsequently used as inputs in all our evaluations while, the IMU measurements were used directly as is without any addition of error terms.

%where \(s_{DVL}=0.007\) is the scale-factor error, \(\mathbf b_{DVL}=0.001\,\mathbf 1_4\) is the bias vector, \(\sigma_{DVL}=0.042\) is the noise standard deviation, and

% \begin{equation}
% \boldsymbol{\epsilon}_k
% \sim
% \mathcal N(\mathbf 0,\mathbf I_4).
% \end{equation}

\subsection{Performance Metrics} \label{metric}

 To evaluate the velocity estimation performance, five quantitative metrics are employed. Let the predicted and ground-truth (GT) velocity vectors at time $t_i$ be denoted by $\hat{\mathbf{v}}(t_i)$ and $\mathbf{v}(t_i)$, respectively.

\begin{enumerate}

\item  Velocity absolute error (VAE):
\begin{equation}
\text{VAE}_i
=
\left\|
\hat{\mathbf{v}}(t_i)-\mathbf{v}(t_i)
\right\|_2.
\end{equation}

\item  Velocity mean absolute error (VMAE):
\begin{equation}
\text{VMAE}
=
\frac{1}{N}
\sum_{i=1}^{N}
\text{VAE}_i,
\end{equation}
where $N$ denotes the total number of samples.

\item  Velocity root mean square error (VRMSE):
\begin{equation}
\text{VRMSE}
=
\sqrt{
\frac{1}{N}
\sum_{i=1}^{N}
(\text{VAE}_i)^2
}.
\end{equation}

\item  Coefficient of determination ($R^2$):
\begin{equation}
R^2(\dot{x}_j,\hat{\dot{x}}_j)
=
1-
\frac{
\sum_{i=1}^{N}
\left(
\dot{x}_{j,i}-\hat{\dot{x}}_{j,i}
\right)^2
}
{
\sum_{i=1}^{N}
\left(
\dot{x}_{j,i}-\bar{\dot{x}}_j
\right)^2
},
\end{equation}
where $\dot{x}_{j,i}$ and $\hat{\dot{x}}_{j,i}$ denote the GT and predicted values of the $j$-th velocity component, respectively, and $\bar{\dot{x}}_j$ is the mean of the corresponding GT component.

\item  Variance accounted for (VAF):
\begin{equation}
\mathrm{VAF}(\dot{x}_j,\hat{\dot{x}}_j)
=
\left[
1-
\frac{
\mathrm{var}
\left(
\dot{x}_j-\hat{\dot{x}}_j
\right)
}
{
\mathrm{var}
\left(
\dot{x}_j
\right)
}
\right]
\times 100.
\end{equation}

\end{enumerate}

%%%%%%%%%%%%%%%%%%%%%%

\subsection{Implementation}
The proposed DVL-DeepONet framework was implemented in Python using the PyTorch deep learning library. All experiments were conducted on the hardware platform described in Table~\ref{tab:gpu_config1}.

%%%%%%%%%%%%%%%%%%%%%%%%%%%

\begin{table}[!ht]
	\centering
	\caption{Hardware configuration for DVL-DeepONet.}
	\label{tab:gpu_config1}
    \footnotesize
    \setlength{\tabcolsep}{15pt}
	\renewcommand{\arraystretch}{0.85}
	\begin{tabular}{ll}
		\toprule
		\textbf{Component} & \textbf{Specification} \\
		\midrule
		GPU              & NVIDIA GeForce RTX 4090 \\
		
		OS            & Linux (Debian) \\
		Architecture      & x86\_64 \\
		
		CUDA Version      & 11.8 \\
		cuDNN Version     & 9.1 \\
		RAM        & 67.26 GB \\
		Memory        & 25.28 GB \\
		Tensor Cores             & 512\\
		CPU Cores         & 24 \\
		
		\bottomrule
	\end{tabular}
\end{table}

%%%%%%%%%%%%%%%%%%%%%%%%%%%

% =========================================================

%%%%%%%%%%%%%%%%%%%%%%%%%

\subsection{Noise-Resilient Estimation} 

Table~\ref{tab:ls_deeponet_1} summarizes the performance of the proposed DVL-DeepONet-I framework under noisy IMU/DVL measurements and compares it against the model-based LS solution~\eqref{eq:1} and learning-based BeamsNetV1~\cite{cohen2022beamsnet} model. 
BeamsNetV1 is a CNN-based velocity reconstruction framework that jointly exploits DVL beam measurements and IMU observations to estimate the AUV velocity vector. Owing to its sensor-fusion capability and demonstrated performance under noisy conditions, it serves as a strong learning-based baseline for comparison.

% =========================================================

\begin{table*}[htbp]
\centering
\caption{Performance comparison for noisy IMU/DVL measurements.}
\label{tab:ls_deeponet_1}
\footnotesize
\begin{tabular}{lccccc}
\toprule
Model &
VRMSE $\downarrow$ &
VMAE $\downarrow$ &
Mean $R^2$ $\uparrow$ &
VAF (\%) $\uparrow$ &
\makecell{VRMSE\\Gain (\%) $\uparrow$} \\
\midrule

Classical LS
& 0.129 & 0.115 & 0.846 & 84 & 18 \\

Vanilla-CNN
& 0.116 & 0.097 & 0.880 & 88 & 10 \\

BeamsNetV1
& 0.123 & 0.110 & 0.867 & 86 & 15 \\

DVL-DeepONet-I
& \textbf{0.105}
& \textbf{0.093}
& \textbf{0.907}
& \textbf{91}
& -- \\
\bottomrule
\end{tabular}
\end{table*}
% =========================================================

%The results demonstrate that the proposed operator-learning approach consistently achieves superior velocity estimation accuracy despite the presence of sensor noise.

The standard LS estimator yields a VRMSE of 0.129~m/s and a VMAE of 0.115~m/s. Although LS estimator provides a direct analytical reconstruction of the vehicle velocity from the DVL beam measurements, its performance deteriorates under noisy conditions. This is because measurement errors are directly propagated through the inversion process. However, learning-based algorithms such as Vanilla-CNN and BeamsNetV1 improve estimation accuracy 
marginally over the model-based algorithm due to their inherent noise-reduction ability.

The proposed LS-DeepONet-I significantly outperforms all baseline methods. It achieves the lowest velocity reconstruction error, with a VRMSE of 0.105~m/s and a VMAE of 0.093~m/s. Compared with the LS estimator, Vanilla-CNN, and BeamsNetV1 approaches, the proposed model yields performance improvements of 18\%, 10\%, and 15\%, respectively, corresponding to an average improvement of approximately 15\%.

%classical LS solution, DVL-DeepONet-I reduces the VRMSE by approximately 18\% and the VMAE by 19\%. Similarly, relative to BeamsNet, the proposed model achieves reductions of 15\% and 16\% in VRMSE and VMAE, respectively.

Beyond average error metrics, DVL-DeepONet-I also demonstrates superior agreement with the GT velocity dynamics, achieving the highest mean $R^2$ score of 0.907 and VAF of 91\%. These results indicate that the proposed operator-learning framework more accurately captures the underlying nonlinear relationship between noisy DVL/IMU measurements and vehicle velocity.

% \textbf{\textit{to be in discussion section}}\\

% The significantly lower VRMSE and VMAE values indicate that DVL-DeepONet-I provides more consistent velocity estimates across the entire trajectory. This observation is further supported by the higher $R^2$ and VAF values, confirming improved robustness to measurement noise.\\

\begin{figure*}[]
\centering

\subfloat[Test Trajectory 4\label{fig:error_std_t4}]{
    \includegraphics[width=0.48\linewidth]{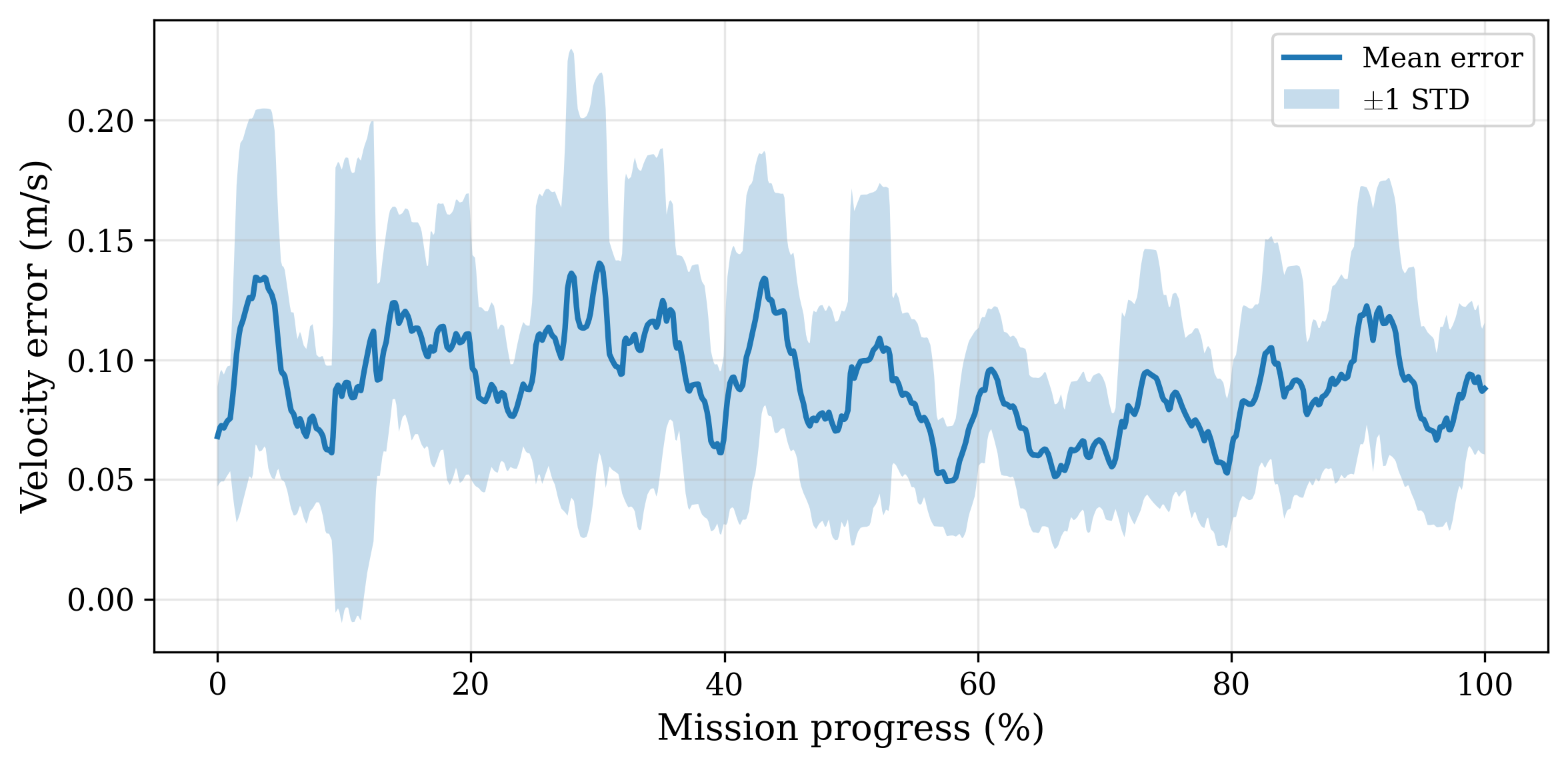}
}
\hfill
\subfloat[Test Trajectory 5\label{fig:error_std_t5}]{
    \includegraphics[width=0.48\linewidth]{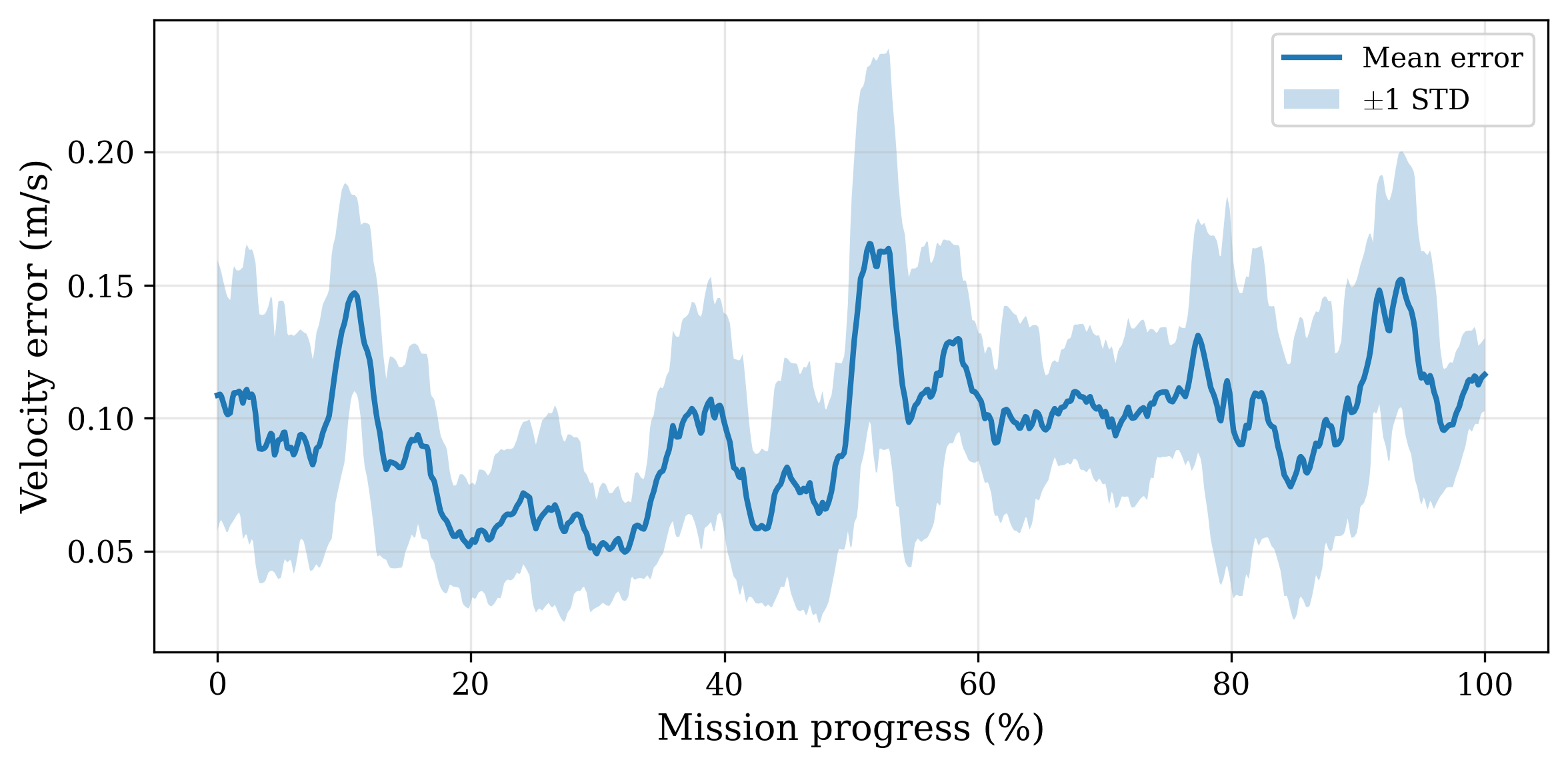}
}

\caption{Velocity reconstruction error profiles of Trajectories 4 and 5 using DVL-DeepONet-I.}
\label{fig:error_std_comparison}
\end{figure*}

Figures~\ref{fig:error_std_t4} and \ref{fig:error_std_t5} illustrate the velocity reconstruction error profiles for the two unseen test trajectories in split 1. For Trajectory 4, the mean velocity error remains relatively consistent throughout the mission, oscillating around 0.08--0.12~m/s.
Trajectory 5 exhibits slightly larger error fluctuations, as the mean error increases to approximately 0.15--0.17~m/s. 
%It is due to the most complex maneuver of the AUV.
However, none of the trajectories exhibits a sustained growth in error over mission progress. This behavior suggests that DVL-DeepONet does not accumulate drift over time and maintains stable performance over long-duration missions. 

%%%%%%%%%%%%%%%%%%%%%%%%%%%%%%%%

%%%%%%%%%%%%%%%%%%%%%%%%%%%%%%%%

%An additional Kalman filtering stage further enhances performance. The DVL-DeepONet+KF model achieves the best overall accuracy with an RMSE of 0.065~m/s and a MAE of 0.060~m/s, corresponding to approximately 47% improvement over Classical LS. The filtering stage effectively suppresses high-frequency prediction fluctuations while preserving the underlying motion dynamics, leading to smoother and more physically consistent velocity estimates.

\subsection{DVL-Only Learning}

To evaluate the DVL-DeepONet-II framework under reduced sensor availability, a second experiment was conducted using only DVL beam measurements, without any IMU information. This scenario is particularly relevant for low-cost underwater platforms and situations where inertial sensors become unavailable or unreliable. In such cases, velocity estimation must rely solely on the information contained in the four DVL beams. For the baseline comparison, we have used model-based LS estimator alongside learning-based vanilla-CNN and BeamsNetV2~\cite{cohen2022beamsnet} models.
BeamsNetV2 is a DVL-only neural baseline that maps four-beam DVL measurements to the velocity vector. The model combines the current DVL beam vector with temporal features derived from a short history of previous beam measurements to predict the AUV velocity components.

Table~\ref{tab:noise_comparison} gives the performance of different baseline models when only DVL measurements are available. The proposed DVL-DeepONet-II achieves substantially the best overall performance with a VRMSE of 0.095~m/s and a VMAE of 0.085~m/s. Compared with the LS estimator, DVL-DeepONet-II reduces both VRMSE and VMAE by approximately 27\%. Furthermore, the proposed method outperforms Vanilla-CNN and BeamsNetV2 baselines, achieving VRMSE reductions of 19\% and 68\%, respectively. Overall, these results correspond to an average improvement of approximately 38\% over the considered baseline methods.

\begin{table*}[htbp]
\centering
\caption{Performance comparison under DVL-only measurements.}
\label{tab:noise_comparison}
\footnotesize
\begin{tabular}{lccccc}
\toprule
Model &
VRMSE $\downarrow$ &
VMAE $\downarrow$ &
Mean $R^2$ $\uparrow$ &
VAF (\%) $\uparrow$ &
\makecell{VRMSE\\Gain (\%) $\uparrow$} \\
\midrule

Classical LS
& 0.129 & 0.115 & 0.846 & 84 & 27 \\

% DNN
% & 0.125 & 0.106 & 0.862 & 86 & 24 \\

Vanilla-CNN
& 0.117 & 0.100 & 0.866 & 87 & 19 \\

BeamsNetV2 
& 0.300 & 0.258 & -0.059 & 78 & 68 \\

DVL-DeepONet-II
& \textbf{0.096}
& \textbf{0.085}
& \textbf{0.905}
& \textbf{91}
& -- \\
\bottomrule
\end{tabular}
\end{table*}

In addition, DVL-DeepONet-II attains the highest mean $R^2$ value of 0.905 and the highest VAF of 91\%, indicating a stronger agreement with the GT velocity dynamics. Although both DNN and Vanilla-CNN improve upon the classical LS solution, their performance remains inferior to that of DVL-DeepONet-II in all reported metrics. The results suggest that purely data-driven architectures are unable to fully exploit the underlying physical relationship between DVL beam measurements and vehicle velocity.

\subsection{Beam Measurement Recovery}
To evaluate the robustness of DVL-DeepONet-III under degraded sensing conditions, artificial beam outages were introduced in both the training and testing datasets. 
Specifically, one or two DVL beam measurements were randomly removed every two seconds, corresponding to every second DVL epoch. The corresponding unavailable beam measurements were masked using the beam-availability matrix defined in \eqref{eq:10} and \eqref{eq:11}, while the synchronized IMU measurements remained available throughout the experiment.

  Then, our proposed algorithm is compared with baseline model-based extended loosely coupled (ELC) approach~\cite{cohen2024seamless}, and with learning-based MissBeamNet~\cite{yona2024missbeamnet}.
%In both model-based baselines, the INS velocity prediction is used as the prior estimate, while the available DVL beams provide measurement updates. In the TC architecture, the available raw DVL beams are directly used in the update step. On the other hand, in the ELC formulation, 
In the ELC approach, the missing beams are generated by virtual beams using the assumption of zero sway velocity or taking the last estimated velocity vector. Both approaches outperformed a simple average on the missing beams history.
For the learning-based baselines, MissBeamNet employs an LSTM network to estimate missing DVL beam measurements from the available beams and their temporal evolution. The reconstructed beam vector is subsequently processed using the LS estimator to obtain the velocity vector, thereby providing a strong learning-based baseline for missing-beam scenarios.

%Unlike model-based approaches that attempt to reconstruct the missing beams prior to velocity estimation, the proposed LS-DeepONet-III directly receives the partially observed DVL beam measurements together with the IMU observations. The branch encoder learns temporal correlations between the available DVL beams and inertial measurements, while the physics-guided loss enforces consistency between the predicted velocity and the available beam observations. Consequently, the network learns to infer the underlying vehicle velocity even when the DVL beam geometry becomes incomplete, thereby improving resilience to beam outages and degraded sensing conditions.}}

 %forcing the model to infer the vehicle velocity from incomplete DVL information aided by inertial measurements.

Table~\ref{tab:partial_beam_comparison} presents the performance of proposed algorithm under partial DVL beam availability. Under this challenging scenario, ELC approach exhibits substantial performance degradation, yielding VRMSE values close to 2~m/s. The corresponding negative $R^2$ and VAF values indicate that the traditional model-based techniques are less effective.
%To ensure a fair and physically meaningful comparison, the estimated velocities from model-based techniques were constrained to the range observed in the GT dataset. Without this constraint, the analytical methods lead to RMSE values above 200~m/s. However, the proposed DVL-DeepONet-III did not require such stabilization and consistently generated bounded velocity estimates.

\begin{table*}[htbp]
\centering
\caption{Performance comparison under partial DVL beam availability.}
\label{tab:partial_beam_comparison}
\footnotesize
\begin{tabular}{lccccc}
\toprule
Model &
VRMSE $\downarrow$ &
VMAE $\downarrow$ &
Mean $R^2$ $\uparrow$ &
VAF (\%) $\uparrow$ &
\makecell{VRMSE\\Gain (\%) $\uparrow$} \\
\midrule

% TC
% & 1.857 & 1.752 & -37.640 & -2246 & 93 \\

ELC
& 1.798 & 1.496 & -38.428 & -929 & 92 \\

MissBeamNet
& 0.183 & 0.153 & 0.601 & 62 & 37 \\

DVL-DeepONet-III
& \textbf{0.108}
& \textbf{0.095}
& \textbf{0.886}
& \textbf{89}
& -- \\
\bottomrule
\end{tabular}
\end{table*}

%This behavior highlights the sensitivity of traditional model-based techniques and the resulting ill-conditioned velocity estimation problem.

MissBeamNet improves the reconstruction accuracy by LSTM algorithm.  Consequently, the VRMSE decreases from approximately 1.8~m/s to 0.182~m/s, while the mean $R^2$ and VAF increase to 0.624 and 64\%, respectively. %These results demonstrate the effectiveness of data-driven beam recovery; however, a noticeable performance gap remains due to error propagation from the beam reconstruction stage to the final velocity estimation.

The proposed DVL-DeepONet-III achieves the best performance across all evaluation metrics, obtaining a VRMSE of 0.114~m/s and a VMAE of 0.099~m/s. Compared with ELC, the proposed method reduces the VRMSE by approximately 92\%, respectively. Furthermore, DVL-DeepONet-III outperforms MissBeamNet by reducing the VRMSE by approximately 37\%. The proposed framework also achieves the highest mean $R^2$ score of 0.873 and VAF of 88\%, indicating a significantly stronger agreement with the GT velocity dynamics.
Overall, DVL-DeepONet-III achieves 65\% improvement over baselines.

\subsection{Ablation Study}
To better understand the contribution of the individual components of the proposed framework, an ablation study was conducted focusing on two aspects of DVL-DeepONet-I (i) temporal observation window length and (ii) cross-validation. 
%The objective was to identify which design choices contribute most significantly to the final performance.

\subsubsection{Effect of Temporal Window Length}

\begin{figure}[]
    \begin{center}
        \includegraphics[height=6cm,width=9cm]{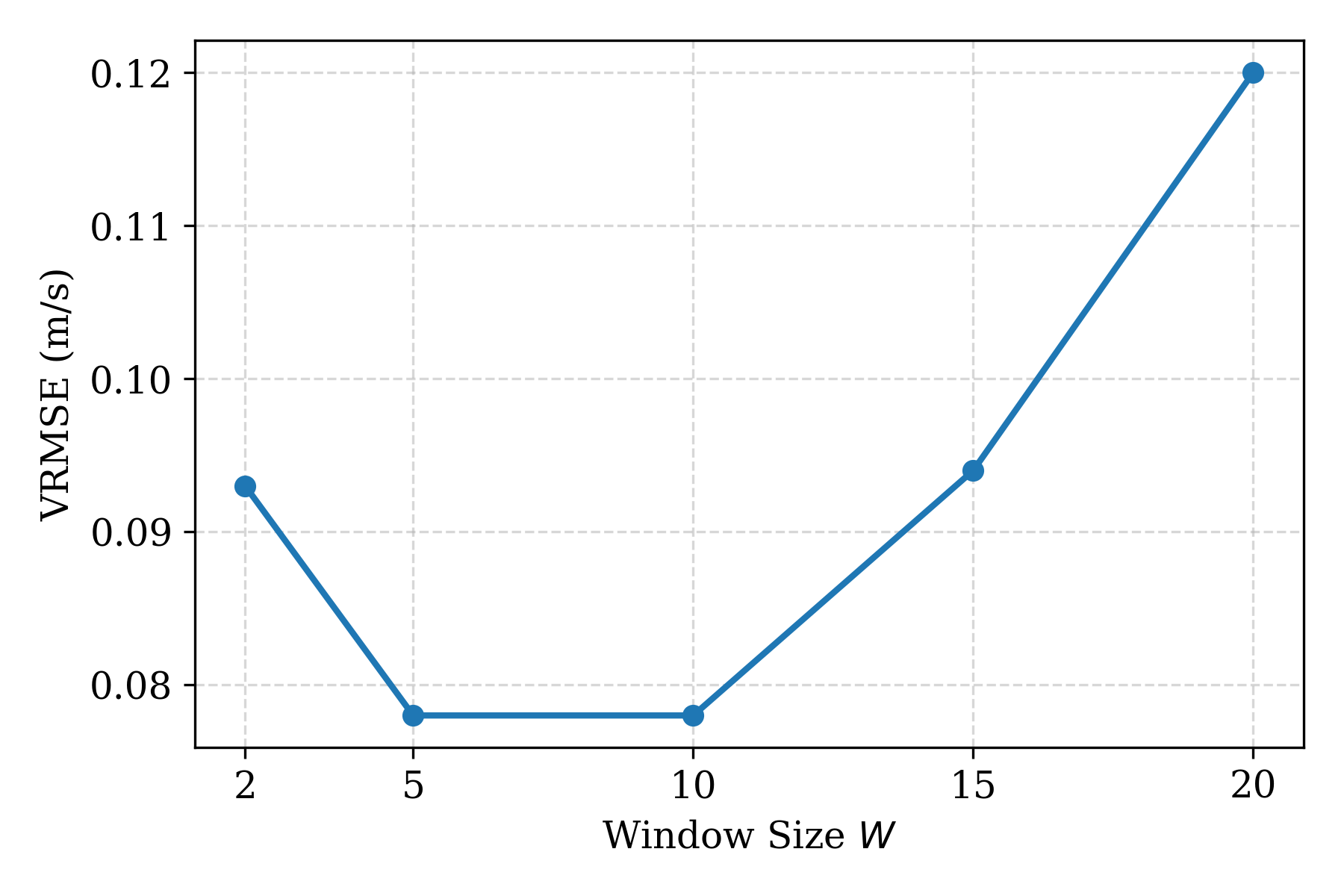}
        \caption{Ablation study of the influence of temporal window size $W$ on the VRMSE of DVL-DeepONet  in the noise-resilient estimation scenario. }
        \label{fig:ablation_window}
    \end{center}
\end{figure}

The first ablation study investigated the influence of the temporal observation window size ($W$) used by the branch network. 
To this end, the model was trained and evaluated using window sizes ranging from $W=2$ to $W=20$, while keeping all other hyperparameters unchanged. The  VRMSE results as a function of the window size are presented in Fig.~\ref{fig:ablation_window}.
As shown in Fig.~\ref{fig:ablation_window}, the estimation accuracy initially improves as the window size increases. The best performance is achieved at $W=10$, yielding a VRMSE of 0.078~m/s. Compared with $W=2$, the VRMSE is reduced by approximately 17\%, demonstrating the benefit of incorporate additional temporal information. However, further increasing the window size beyond $W=10$ leads to a gradual increase in VRMSE.
This behavior suggests that excessively long windows may introduce redundant or less informative historical measurements.

 %Similarly, in the DVL-only scenario, increasing the window length further to 15 samples reduced the RMSE to 0.082~m/s. These results indicate that temporal information plays a crucial role in underwater velocity estimation, as the network can exploit motion continuity and sensor correlations over longer periods.

%%%%%%%%%%%%%%%

%%%%%%%%%%%%%%%

% A noteworthy observation from the ablation studies is that increasing the temporal observation window from 2 to 10 samples substantially improved performance. This suggests that, for noisy DVL/IMU fusion, access to a longer temporal history is more valuable than increasing feature dimensionality. Consequently, the final DVL-DeepONet-I architecture employs raw normalized sensor measurements together with an extended temporal window.

% Overall, the results demonstrate that the proposed DVL-DeepONet framework effectively combines data-driven operator learning with physical consistency constraints, yielding accurate and robust velocity estimation under realistic noisy underwater navigation conditions.

% A key finding of the study is that extending the temporal observation window substantially improves performance, highlighting the importance of temporal operator learning for underwater navigation. 

%%%%%%%%%%%%

\subsubsection{Cross-Validation Study} \label{sec:cross}

\begin{figure*}[]
\centering

\subfloat[Trajectory 2]{
\includegraphics[width=0.32\textwidth]{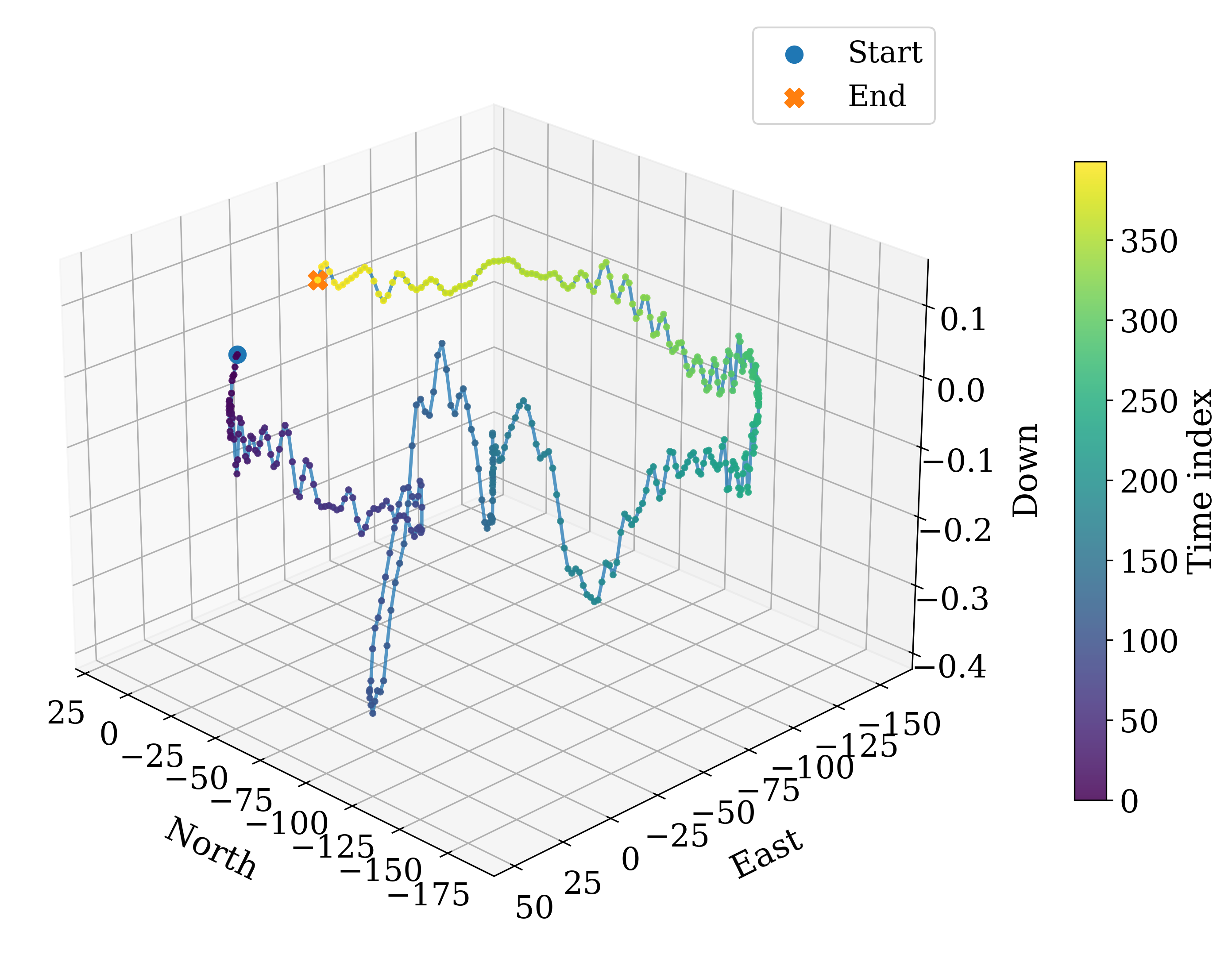}
}
\subfloat[Trajectory 3]{
\includegraphics[width=0.32\textwidth]{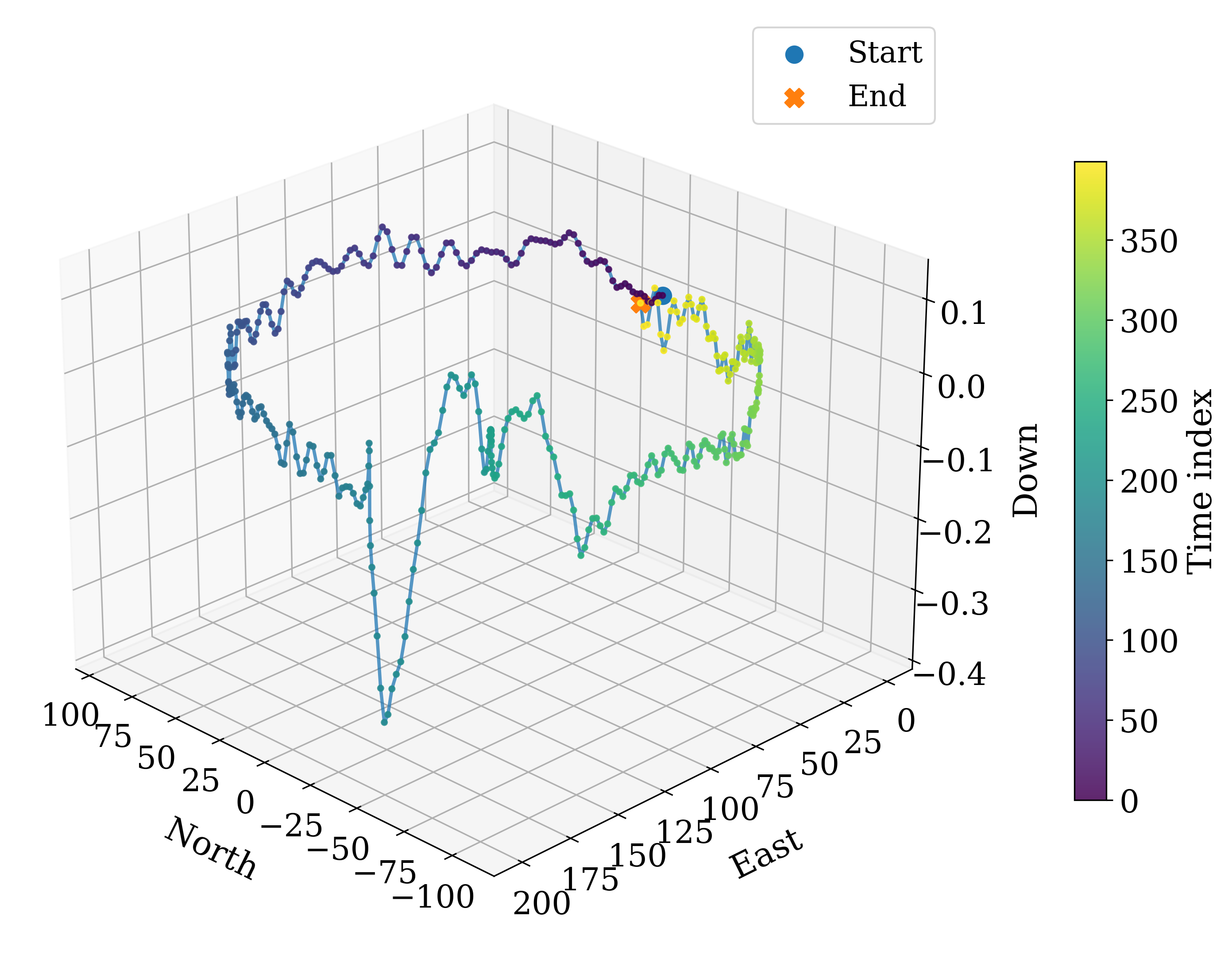}
}
\subfloat[Trajectory 4]{
\includegraphics[width=0.32\textwidth]{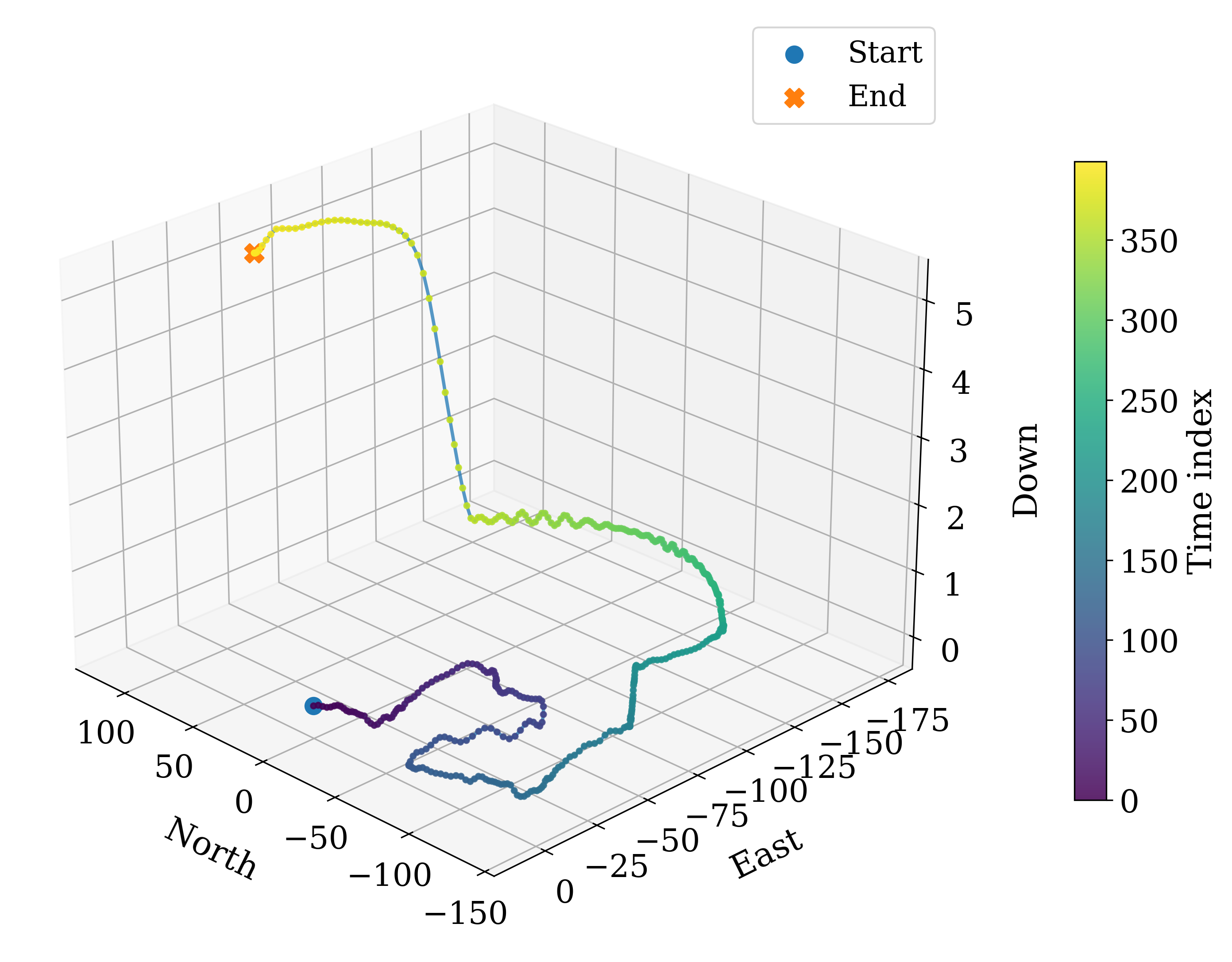}
}

\vspace{0.15cm}

\subfloat[Trajectory 5]{
\includegraphics[width=0.32\textwidth]{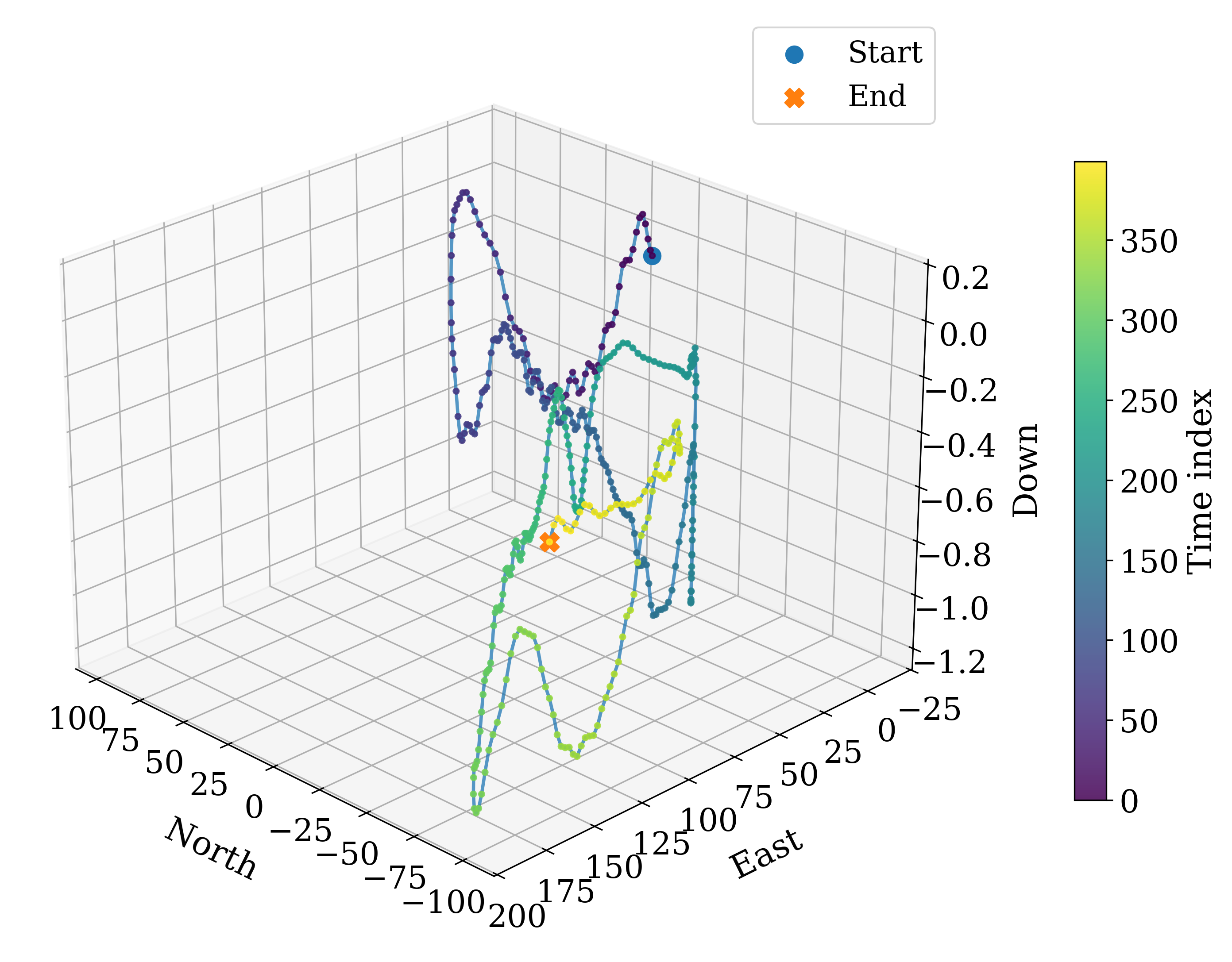}
}
\subfloat[Trajectory 9]{
\includegraphics[width=0.32\textwidth]{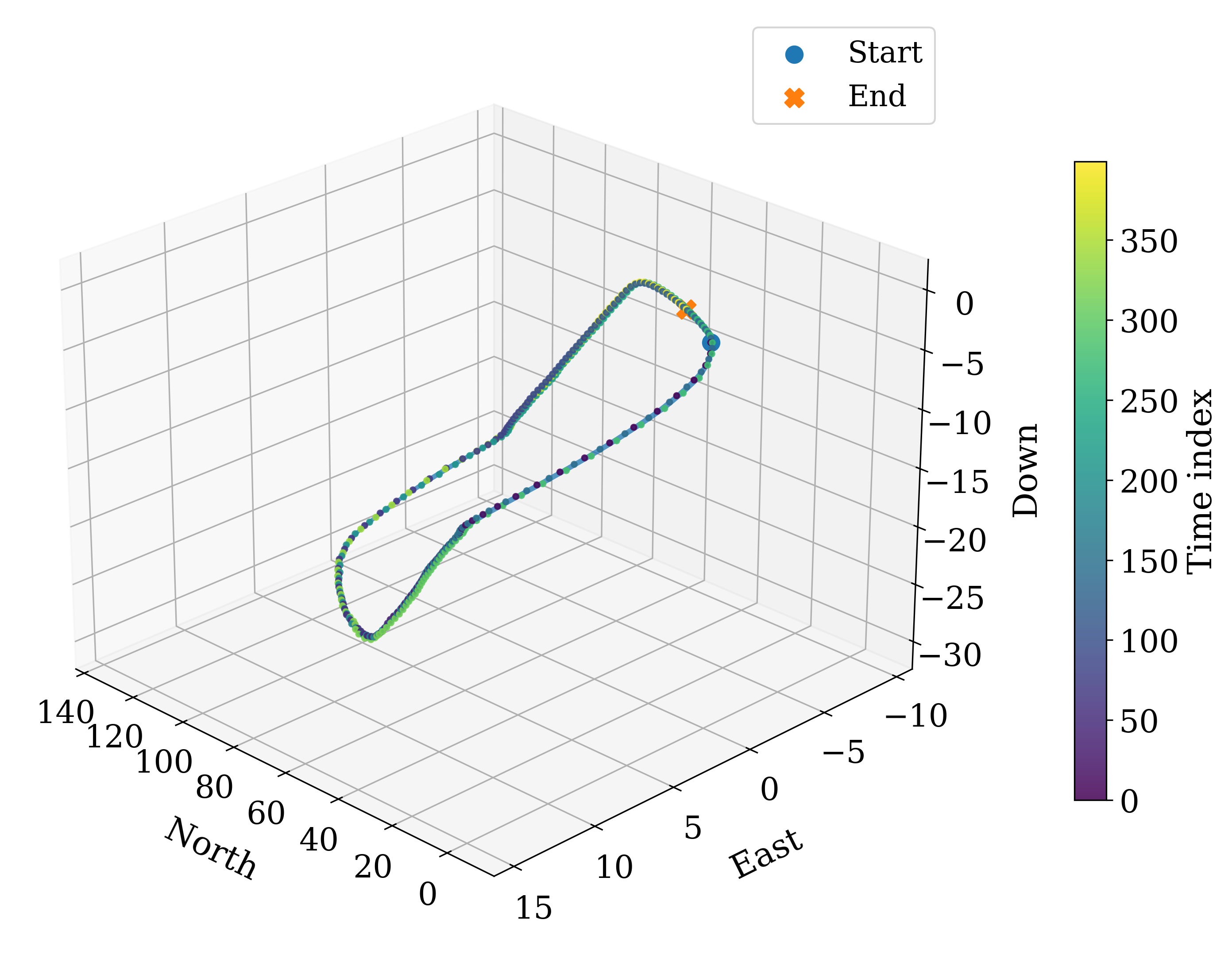}
}
\subfloat[Trajectory 10]{
\includegraphics[width=0.32\textwidth]{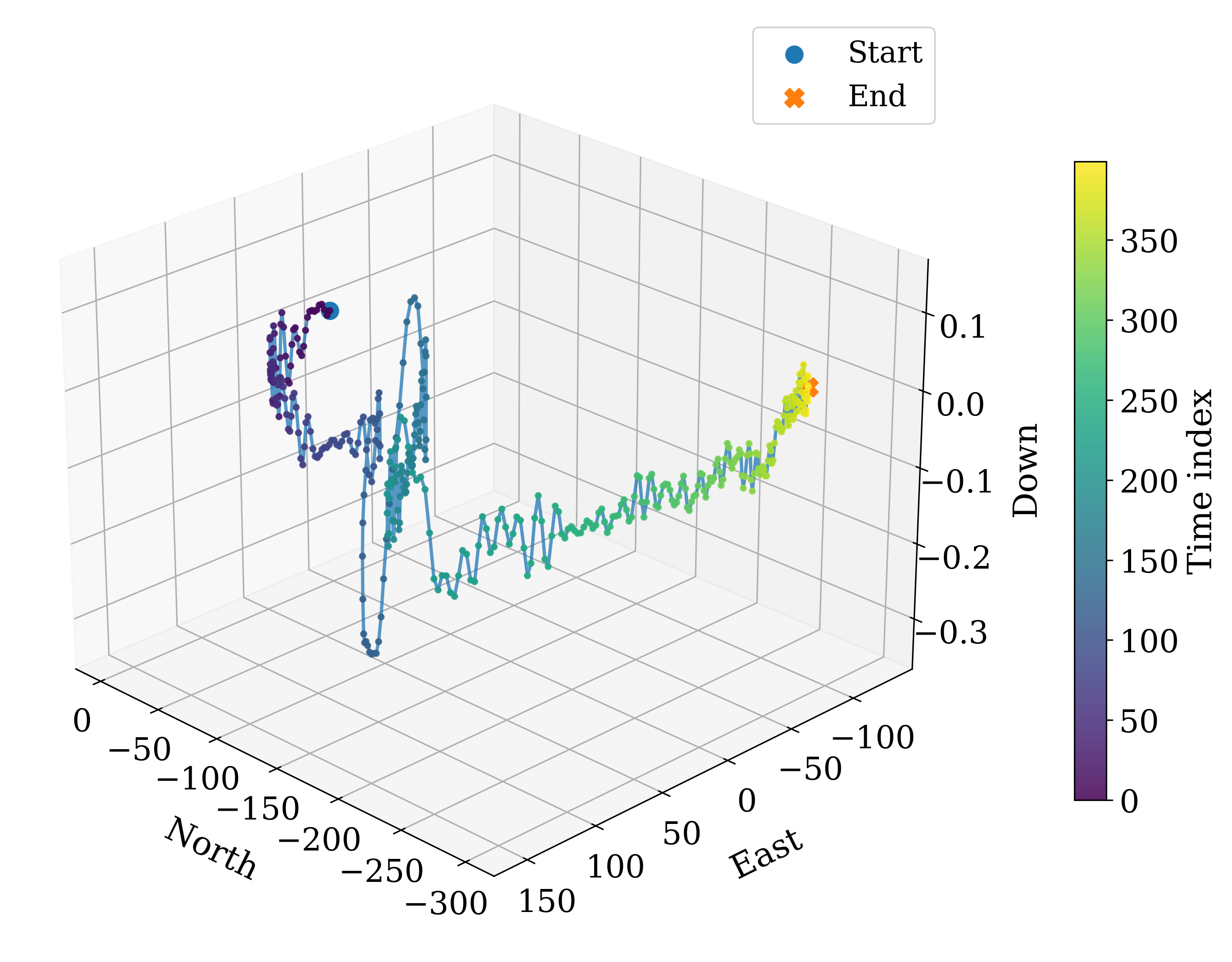}
}

\caption{GT trajectories used as unseen test missions in the three-fold cross-validation study. }
\label{fig:test_trajectories}
\end{figure*}

Table~\ref{tab:cv_splits} summarizes the trajectory distribution used in the three-fold cross-validation study. In each fold, nine trajectories are used for training, two trajectories for validation, and two for testing, corresponding to an approximate 70\%-15\%-15\% train-validation-test split. Fig.~\ref{fig:test_trajectories} shows the GT trajectories used as unseen test missions in the three-fold cross-validation study. 
Table~\ref{tab:cross_validation_results} incorporates cross-validation results of the proposed DVL-DeepONet-I framework in the noise-resilient estimation scenario. While BeamsNet provides only marginal gains of 5\% on average, over the LS estimator, DVL-DeepONet-I achieves substantially larger improvements of 21\%. 
The consistent gains across all splits demonstrate that the proposed model is a reliable solution for practical underwater navigation scenarios.

%The proposed operator-learning framework consistently achieves approximately four to six times larger gains than BeamsNetV1. This suggests that incorporating the physical structure of the velocity reconstruction problem through the DeepONet architecture leads to substantially better generalization than conventional neural-network approaches.

\begin{table*}[htbp]
\centering
\caption{Cross-validation splits used for the evaluations.}
\label{tab:cv_splits}
\footnotesize
\setlength{\tabcolsep}{2pt}
\begin{tabular}{ccccccc}
\toprule
Split &
Train. Traj. &
Val. Traj. &
Test Traj. &
Train Dist. (m) &
Val. Dist. (m) &
Test Dist. (m) \\
\midrule

S1 &
1,6,7,8,9,10,11,12,13 &
2,3 &
4,5 &
7094.26 &

1346.59 &
1566.93 \\

S2 &
4,5,6,7,8,9,10,11,12 &
1,13 &
2,3 &
7259.75 &
1496.86 &
1346.59 \\

S3 &
1,2,3,4,5,6,11,12,13 &
7,8 &
9,10 &
6712.22 &
1686.80 &
1608.74 \\

\bottomrule
\end{tabular}
\end{table*}
% =========================================================

%%%%%%%%%%%%%%%%%%%%%%%%%%%%%%%%
\begin{table*}[!ht]
\centering
\caption{Cross-validation results for the noise-resilient estimation scenario.}
\label{tab:cross_validation_results}
\footnotesize
\begin{tabular}{cccccc}
\toprule
Split & Method & VRMSE$_x$ & VRMSE$_y$ & VRMSE$_z$ & Improvement over LS (\%) \\
\midrule

\multirow{3}{*}{S1}
& Classical LS & 0.088 & 0.090 & 0.023 & N/A \\
& BeamsNetV1 & 0.082 & 0.088 & 0.021 & 05 \\
& \textbf{DVL-DeepONet-I } & \textbf{0.070} & \textbf{0.077} & \textbf{0.017} & \textbf{18} \\
\midrule

\multirow{3}{*}{S2}
& Classical LS & 0.088 & 0.089 & 0.023 & N/A \\
& BeamsNetV1 & 0.084 & 0.084 & 0.021 & 04 \\
& \textbf{DVL-DeepONet-I } & \textbf{0.060} & \textbf{0.062} & \textbf{0.012} & \textbf{31} \\
\midrule

\multirow{3}{*}{S3}
& Classical LS & 0.086 & 0.087 & 0.024 & N/A \\
& BeamsNetV1 & 0.081 & 0.082 & 0.021 & 06 \\
& \textbf{DVL-DeepONet-I } & \textbf{0.071} & \textbf{0.088} & \textbf{0.011} & \textbf{14} \\
\midrule

\multicolumn{5}{r}{\textbf{Avg. Improvement }} &
\textbf{21} \\
\bottomrule
\end{tabular}
\end{table*}
%%%%%%%%%%%%%%%%%%%%%%%%%%%%%%%%

\subsection{Summary}

%The experimental results demonstrate the effectiveness and robustness of the proposed DVL-DeepONet framework under three representative underwater navigation scenarios: noisy IMU/DVL measurements, DVL-only operation, and partial IMU/DVL availability. In the noisy sensor case, DVL-DeepONet-I significantly outperformed both Classical LS and BeamsNetV1, achieving up to 47\% improvement in VRMSE. In the DVL-only scenario, the proposed method maintained strong estimation capability despite the absence of inertial information, reducing the  VRMSE by approximately 33\% compared with the classical LS solution. Furthermore, under partial sensor availability, the framework successfully leveraged temporal correlations and complementary sensor information to provide reliable velocity estimates despite missing measurements and achieved an average of 70\% improvements.

The proposed DVL-DeepONet framework was evaluated in three challenging underwater navigation scenarios 
and demonstrated supervisor performance against other model-based and data-driven approaches as shown in Table~\ref{tab:dvl_deeponet_summary}.

\begin{table}[htbp]
\centering
\caption{Summary of the three DVL-DeepONet configurations with the best VRMSE improvement achieved in each scenario.}
\label{tab:dvl_deeponet_summary}
\footnotesize
\begin{tabular}{lcc}
\toprule

\textbf{Scenario} &
\textbf{VRMSE} $\downarrow$ &
\textbf{Best Gain (\%)} $\uparrow$ \\
\midrule

 Noise-Resilient Estimation
& 0.105
& 18 \\

 DVL-Only Learning
& 0.096
& 68 \\

 Beams Measurement Recovery
& 0.114
& 92 \\

\bottomrule
\end{tabular}
\end{table}

For the noisy IMU/DVL scenario, DVL-DeepONet-I achieved average performance improvements of 15\% over the baselines. Similarly, in the DVL-only scenario, the average gain by LS-DeepONet-II is 38\%. Finally, under the measurement recovery, our model registered 65\% improvement. Moreover, across all scenarios, the proposed operator-learning framework consistently outperformed both model-based approaches and conventional deep-learning baselines. Additionally, it maintains around 90\% agreement with the GT, as evidenced by mean $R^2$ values.

\section{Conclusion} \label{sec:6}

Existing DVL-based velocity estimation methods either rely on model-based LS estimators or employ purely data-driven models. Moreover, the model-based algorithms become unreliable under beam outages and rank-deficient measurement configurations, and data-driven models are black-box in nature and do not explicitly enforce the physical relationship between DVL beam measurements and AUV velocity.
To address these limitations, this paper proposed DVL-DeepONet, a physics-guided operator-learning framework for velocity estimation of AUV. The proposed method learns a nonlinear mapping from inertial/DVL observations to vehicle velocity and incorporates the DVL observation model as a physics-based consistency constraint during training.
 Unlike conventional LS reconstruction, the framework does not require direct inversion of the beam geometry matrix and therefore remains applicable even in beam outrages scenarios. Furthermore, the integration of physical constraints improves interpretability and preserves consistency between the predicted velocity vector and the measured DVL beams. As a result, DVL-DeepONet provides robust velocity estimation under noisy measurements, partial beam availability, and degraded beam conditions.

The proposed framework has been evaluated using approximately 10,000~m of real-world AUV experimental data collected during multiple sea trials.
Experimental results demonstrated that, DVL-DeepONet-I achieved average performance improvements of approximately 15\% under noisy inertial/DVL measurements, DVL-DeepONet-II achieved improvements of about 38\% in the DVL-only scenario, and DVL-DeepONet-III gains exceeding 65\% under partial measurement availability. Overall, the proposed framework achieves an average improvement of approximately 40\% compared with the baseline methods.

These results demonstrate that the proposed operator-learning frameworks successfully handles multiple practical operational scenarios using a unified architecture. The study further shows that integrating physics constraints within DeepONet architectures improves accuracy, robustness and generalization.  Moreover, the results indicate that physically informed operator learning can serve as a viable alternative to purely model-based and purely data-driven approaches for underwater navigation tasks.

%In the noisy IMU/DVL setting, the DVL-DeepONet-I achieves approximately 15\% higher accuracy than baseline approaches. When only DVL measurements are available, DVL-DeepONet-II maintains strong velocity estimation performance, making it particularly suitable for low-cost and sensor-constrained underwater platforms, with an improvement of 38\% over the baselines.  Furthermore, under partial sensor availability, the proposed method remains robust to missing DVL beams and incomplete sensor information. In this challenging scenario, DVL-DeepONet-III achieves a remarkable 74\% performance gain. This results in an average performance improvement of 42\% over baselines by our DVL-DeepONet.

Nonetheless, several limitations warrant consideration. First, DVL-DeepONet requires a minimum temporal window size of two samples, whereas the model-based LS estimator can operate using only the current measurement.
 Second, the training stage involves multiple CNN and MLP components, which require GPU acceleration or higher computational power. However, the trained model remains lightweight during inference and is suitable for real-time deployment on AUVs with limited onboard computational resources.

DVL-DeepONet offers a white-box solution for resilient underwater navigation, improving mission safety, and operational effectiveness under degraded sensing conditions.
Since the imposed physical constraints originate from the DVL measurement process itself and are independent of a particular operating region, the proposed framework is expected to generalize more effectively across diverse underwater environments than purely data-driven approaches.
Future work will focus on extending the proposed framework to full navigation-state estimation (position, velocity, and orientation) and integrating uncertainty-aware learning strategies.

\vspace{0.2cm}

\section*{Conflict of Interest Statement}
\noindent The authors confirm that they have no conflicts of interest related to this paper.

\section*{Funding Declaration}
\noindent The authors confirm that they did not receive any funding to carry out this work.

\section*{Data Availability}
\noindent The dataset used in this study is publicly available at: 
\url{https://github.com/ansfl/A-KIT/tree/main}.

\section*{Code Availability}

\noindent The source code associated with this study is publicly available at: \url{https://github.com/ansfl/DVL-DeepONet}.
%%

% Uncomment and use as the case may be
%\begin{theorem} 
%\end{theorem}

% Uncomment and use as the case may be
%\begin{lemma} 
%\end{lemma}


\printcredits

%\bibliographystyle{cas-model2-names}
\bibliographystyle{elsarticle-num}
%\bibliographystyle{plainnat}
\bibliography{Refrences.bib}

@article{DAMARI2026125277,
title = {{ResAlignNet}: {A} data-driven approach for {INS/DVL} alignment},
author = {Guy Damari and Itzik Klein},
journal = {Ocean Engineering},
volume = {356},
pages = {125277},
year = {2026},
issn = {0029-8018},
}

@article{zhang2022submarine,
  title={Submarine pipeline tracking technology based on {AUVs} with forward looking sonar},
  author={Zhang, Yupeng and Zhang, Hongwei and Liu, Jun and Zhang, Shitong and Liu, Zhi and Lyu, Enmou and Chen, Weiyu},
  journal={Applied Ocean Research},
  volume={122},
  pages={103128},
  year={2022},
  publisher={Elsevier}
}

@article{cheng2026robust,
  title={A robust {INS/USBL/DVL} integrated navigation method based on adaptive correlation entropy factor graph optimization},
  author={Cheng, Sen and Wang, Yanyan and Zhao, Qinglong and Zhu, Haitao and Qu, Xianqiang},
  journal={Ocean Engineering},
  volume={356},
  pages={125234},
  year={2026},
  publisher={Elsevier}
}

@article{braginsky2020correction,
  title={Correction of {DVL} error caused by seafloor gradient},
  author={Braginsky, Boris and Baruch, Alon and Guterman, Hugo},
  journal={IEEE Sensors Journal},
  volume={20},
  number={19},
  pages={11652--11659},
  year={2020},
  publisher={IEEE}
}

@article{miao2026physics,
  title={Physics-guided adaptive {UKF} for robust {AUV} integrated navigation under degraded underwater observations},
  author={Miao, Yunhong and Liu, Xiaojie and Sun, Yu and Liu, Xiaochen and Shen, Chong and Wang, Chenguang and Tang, Jun and Liu, Jun},
  journal={Ocean Engineering},
  volume={362},
  pages={126542},
  year={2026},
  publisher={Elsevier}
}

@article{wang2019novel,
  title={A novel {SINS/DVL} tightly integrated navigation method for complex environment},
  author={Wang, Di and Xu, Xiaosu and Yao, Yiqing and Zhang, Tao and Zhu, Yongyun},
  journal={IEEE Transactions on Instrumentation and Measurement},
  volume={69},
  number={7},
  pages={5183--5196},
  year={2019},
  publisher={IEEE}
}

@article{mo2026hybrid,
  title={A Hybrid Physics--Data-Driven Navigation Method for {AUVs} Fusing Hydrographic Information with {INS/DVL} Integration},
  author={Mo, Haozhou and Yang, Hongze and Zhang, Yanbei and Pan, Daqing and Yang, Gongliu and Li, Wenqiang},
  journal={IEEE Sensors Journal},
  year={2026},
  publisher={IEEE}
}

@article{kang2026hybrid,
  title={A hybrid-kernel-based adaptive robust {Kalman} filter for {INS/DVL} integrated underwater navigation},
  author={Kang, Le and He, Kunpeng and Zhao, Jinyue and Wang, Xiangdong and Tan, Panlong},
  journal={Ocean Engineering},
  volume={350},
  pages={124269},
  year={2026},
  publisher={Elsevier}
}

@article{engelsman2023information,
  title={Information-aided inertial navigation: {A} review},
  author={Engelsman, Daniel and Klein, Itzik},
  journal={IEEE Transactions on Instrumentation and Measurement},
  volume={72},
  pages={1--18},
  year={2023},
  publisher={IEEE}
}

@inproceedings{cohen2022libeamsnet,
  title={{LiBeamsNet}: {AUV} velocity vector estimation in situations of limited {DVL} beam measurements},
  author={Cohen, Nadav and Klein, Itzik},
  booktitle={OCEANS 2022, Hampton Roads},
  pages={1--5},
  year={2022},
  organization={IEEE}
}

@article{DeepONet,
  author = {Lu, Lu and Jin, Pengzhan and Pang, Guofei and Zhang, Zhiping and Karniadakis, George E.},
  title = {Learning nonlinear operators via {DeepONet} based on the universal approximation theorem of operators},
  journal = {Nature Machine Intelligence},
  year = {2021},
  volume = {3},
  pages = {218--229}
}

@article{yona2024missbeamnet,
  title={{MissBeamNet}: Learning missing {Doppler} velocity log beam measurements},
  author={Yona, Mor and Klein, Itzik},
  journal={Neural Computing and Applications},
  volume={36},
  number={9},
  pages={4947--4958},
  year={2024},
  publisher={Springer}
}

@article{paull2013auv,
  title={{AUV} navigation and localization: {A} review},
  author={Paull, Liam and Saeedi, Sajad and Seto, Mae and Li, Howard},
  journal={IEEE Journal of Oceanic Engineering},
  volume={39},
  number={1},
  pages={131--149},
  year={2013},
  publisher={IEEE}
}

@book{wadoo2017autonomous,
  title={Autonomous underwater vehicles: modeling, control design and simulation},
  author={Wadoo, Sabiha},
  year={2017},
  publisher={CRC press}
}

@article{zhang2026novel,
  title={Novel algorithm for the calibration of {DVL} in underwater integrated navigation system},
  author={Zhang, Haixu and Li, Chong and Zhang, Tao and Wang, Guangcai and Wang, Di},
  journal={Ocean Engineering},
  volume={353},
  pages={124676},
  year={2026},
  publisher={Elsevier}
}

@article{zhang2025underwater,
  title={Underwater {DVL} Optimization Network {(UDON)}: A Learning-Based {DVL} Velocity Optimizing Method for Underwater Navigation},
  author={Zhang, Feihu and Zhao, Shaoping and Li, Lu and Cao, Chun},
  journal={Drones},
  volume={9},
  number={1},
  pages={56},
  year={2025},
  publisher={MDPI}
}

@article{batovs2026dmian,
  title={{DMIAN}: deep learning-based Multi-{IMU} fusion for enhanced marine aided navigation},
  author={Bato{\v{s}}, Matko and Na{\dj}, {\DH}ula},
  journal={Control engineering practice},
  volume={173},
  pages={106991},
  year={2026},
  publisher={Elsevier}
}

@article{cohen2024seamless,
  title={Seamless underwater navigation with limited {Doppler} velocity log measurements},
  author={Cohen, Nadav and Klein, Itzik},
  journal={IEEE Transactions on Intelligent Vehicles},
  year={2024},
  publisher={IEEE}
}

@article{mu2019end,
  title={End-to-end navigation for autonomous underwater vehicle with hybrid recurrent neural networks},
  author={Mu, Xiaokai and He, Bo and Zhang, Xin and Song, Yan and Shen, Yue and Feng, Chen},
  journal={Ocean Engineering},
  volume={194},
  pages={106602},
  year={2019},
  publisher={Elsevier}
}

@article{tan2025modeling,
  title={Modeling vehicle dynamics with physics-informed deep operator network},
  author={Tan, Chenkai and Cai, Yingfeng and Wang, Hai and Chen, Long and Lian, Yubo},
  journal={Vehicle System Dynamics},
  pages={1--29},
  year={2025},
  publisher={Taylor \& Francis}
}

@inproceedings{brokloff1994matrix,
  title={Matrix algorithm for {Doppler} sonar navigation},
  author={Brokloff, Ned A},
  booktitle={Proceedings of OCEANS'94},
  volume={3},
  pages={III--378},
  year={1994},
  organization={IEEE}
}

@article{liu2018ins,
  title={{INS/DVL/PS} tightly coupled underwater navigation method with limited {DVL} measurements},
  author={Liu, Peijia and Wang, Bo and Deng, Zhihong and Fu, Mengyin},
  journal={IEEE Sensors Journal},
  volume={18},
  number={7},
  pages={2994--3002},
  year={2018},
  publisher={IEEE}
}

@article{zhang2023autonomous,
  title={Autonomous underwater vehicle navigation: {A} review},
  author={Zhang, Bingbing and Ji, Daxiong and Liu, Shuo and Zhu, Xinke and Xu, Wen},
  journal={Ocean Engineering},
  volume={273},
  pages={113861},
  year={2023},
  publisher={Elsevier}
}

@article{sahoo2026pidr,
  title={{PiDR}: Physics-Informed Inertial Dead Reckoning for Autonomous Platforms},
  author={Sahoo, Arup Kumar and Klein, Itzik},
  journal={arXiv preprint arXiv:2601.03040},
  year={2026}
}

@misc{teledyne_dvl_2023,
  author       = {{Teledyne Marine}},
  title        = {{Doppler Velocity Logs}},
  year         = {2023},
  howpublished = {\url{https://www.teledynemarine.com/products/product-line/navigation-positioning/{Doppler}-velocity-logs}},
  note         = {Accessed: Dec. 2025}
}

@article{shurin2022autonomous,
  title={The autonomous platforms inertial dataset},
  author={Shurin, Artur and Saraev, Alex and Yona, Mor and Gutnik, Yevgeni and Faber, Sharon and Etzion, Aviad and Klein, Itzik},
  journal={IEEE Access},
  volume={10},
  pages={10191--10201},
  year={2022},
  publisher={IEEE}
}

@misc{ixblue_phins_2023,
  author       = {{iXblue}},
  title        = {{PHINS Subsea}},
  year         = {2023},
  howpublished = {\url{https://www.ixblue.com/store/phins-subsea/}},
  note         = {Accessed: Dec. 2025}
}

@misc{eca_a18d_2023,
  author       = {{ECA Group}},
  title        = {{A18-D AUV: Autonomous Underwater Vehicle}},
  year         = {2023},
  howpublished = {\url{https://www.ecagroup.com/en/solutions/a18-d-auv-autonomous-underwater-vehicle}},
  note         = {Accessed: Dec. 2025}
}

@book{chakraverty2025artificial,
	title={Artficial Neural Networks and Type-2 Fuzzy Set: Elements of Soft Computing and Its Applications},
	author={Chakraverty, Snehashish and Sahoo, Arup Kumar  and Mohapatra, D},
	year={2025},
	publisher={Elsevier}
}

@article{cohen2022beamsnet,
	title={{BeamsNet}: A data-driven approach enhancing {Doppler} velocity log measurements for autonomous underwater vehicle navigation},
	author={Cohen, Nadav and Klein, Itzik},
	journal={Engineering Applications of Artificial Intelligence},
	volume={114},
	pages={105216},
	year={2022},
	publisher={Elsevier}
}

@book{titterton2004strapdown,
	title={Strapdown inertial navigation technology},
	author={Titterton, David and Weston, John L},
	volume={17},
	year={2004},
	publisher={IET}
}

@book{groves2013book,
	title={Principles of {GNSS}, Inertial and Multi-Sensor Integrated Navigation Systems},
	author={Groves, P},
	year={2013},
	publisher={Artech House, UK}
}

@book{farrell2007gnss,
	title={GNSS aided navigation \& tracking: inertially augmented or autonomous},
	author={Farrell, James L},
	year={2007},
	publisher={American Literary Press Baltimore, Maryland}
}

@article{yampolsky2025dcnet,
	title={{DCN}et: A data-driven framework for {DVL} calibration},
	author={Yampolsky, Zeev and Klein, Itzik},
	journal={Applied Ocean Research},
	volume={158},
	pages={104525},
	year={2025},
	publisher={Elsevier}
}



\end{document}